\documentclass[lettersize,journal]{IEEEtran}
\usepackage{amsmath,amsfonts,amssymb,amstext}
\usepackage[ruled,linesnumbered]{algorithm2e}
\usepackage{array}
\usepackage[caption=false,font=normalsize,labelfont=sf,textfont=sf]{subfig}
\usepackage{textcomp}
\usepackage{stfloats}
\usepackage{url}
\usepackage{verbatim}
\usepackage{graphicx}
\usepackage{cite}
\usepackage{xcolor}
\usepackage{bm}
\usepackage{booktabs}
\usepackage{multirow}
\usepackage{makecell}
\usepackage{diagbox}
\usepackage[switch]{lineno}
\begin{document}

\title{Online Camera-Pose-Free Stereo Endoscopic Tissue Deformation Recovery With Tissue-Invariant Vision-Biomechanics Consistency}



\author{
Jiahe Chen~\IEEEmembership{Member,~IEEE}, Naoki Tomii,~\IEEEmembership{Member,~IEEE}, Ichiro Sakuma,~\IEEEmembership{Member,~IEEE}, Etsuko Kobayashi~\IEEEmembership{Member,~IEEE}
\thanks{This work was supported by the JST Moonshot R\&D Grant Number JPMJMS2214, the Grant-in-Aid for JSPS Fellows (Number 25KJ1017), and the Kayamori Foundation of Informational Science Advancement. (Corresponding author: Naoki Tomii)}
\thanks{The authors are with the School of Engineering, The University of Tokyo, 7-3-1 Hongo, Tokyo, 113-8656, Japan. Ichiro Sakuma is also with the Center for Research and Collaboration, Tokyo Denki University, 5 Senju Asahi-cho, Adachi-ku, Tokyo, 120-8551, Japan. (email: chenjh@bmpe.t.u-tokyo.ac.jp; naoki\_tomii@bmpe.t.u-tokyo.ac.jp; sakuma@bmpe.t.u-tokyo.ac.jp or i.sakuma@mail.dendai.ac.jp; etsuko@bmpe.t.u-tokyo.ac.jp)}
}

\markboth{Journal of \LaTeX\ Class Files,~Vol.~14, No.~8, August~2021}%
{Shell \MakeLowercase{\textit{et al.}}: A Sample Article Using IEEEtran.cls for IEEE Journals}


\maketitle

\begin{abstract}
Tissue deformation recovery based on stereo endoscopic images benefits surgical navigation and autonomous soft tissue manipulation, yet existing methods often struggle in dynamic, deformable intraoperative environments because they rely on assumptions like quasi-static conditions, accurate camera pose estimation, offline optimization, or tissue-specific biomechanical priors. This work introduces an online, stereo vision-based tissue deformation recovery method equipped with several novel mechanisms to overcome these limitations. Our proposed approach operates in a camera-centric setting, eliminating reliance on camera pose and its associated errors. It represents complex tissue geometry and deformation using high-dimensional mappings and formulates inter-frame recovery as a functional optimization problem with vision-biomechanics-driven constraints. Experiments on in vivo and ex vivo laparoscopic datasets demonstrate robust modeling even with occlusions and irregular camera motion. The method achieves high 3D reconstruction accuracy (0.37±0.27 mm surface distance) and enables dense surface strain estimation, offering a new modality for intraoperative mechanical analysis.
\end{abstract}


\section{Introduction}
\IEEEPARstart{T}{issue} deformation is the change in its 3D structure and connectivity in time. In surgery, tissue deformation is typically caused by tool-tissue interaction and cardio-respiratory pulsation. Analyzing tissue deformation benefits many downstream tasks, such as surgical navigation~\cite{schneider_performance_2021}, surgical simulation~\cite{zhu_real-time_2022}, analyzing tool-tissue interaction~\cite{chen_occlusion-robust_2023, chen_trans-window_2024}, manipulation risk assessment~\cite{yamamoto_tissue_2023}, and autonomous soft tissue manipulation~\cite{hu_occlusion-robust_2024, chen_surgem_2024}.

Deformation recovery is to estimate and quantify the dynamic change of tissue shape, which is typically 4D data involving spatiotemporal information. Most of the existing deformation recovery methods are based on the stereo vision system, such as the binocular endoscopic~\cite{chavarrias_solano_multi-task_2025}, the binocular laparoscopic~\cite{chen_occlusion-robust_2023}, or the binocular microscopic camera~\cite{giannarou_vision-based_2016}. There are typically two subtasks of deformation recovery: 3D reconstruction and inter-frame alignment. 3D reconstruction is to acquire tissue 3D structure, while inter-frame alignment is to establish time domain connection.

Difficulties in tissue deformation recovery are mostly derived from its non-rigid nature. The motion of a rigid object can be completely represented by a 6-degree of freedom (DoF) transformation, while the deformation of a non-rigid object has infinite DoF. Inter-frame alignment of a rigid scene is almost equal to finding the true camera pose in different frames, while the estimated camera pose is less reliable in a deformable scene and less useful for the alignment. On the other hand, the common interference during surgery also hinders accurate deformation estimation. The interference resulted by occlusion, camera motion, specular reflection, lowly textured surface, and surface discontinuity are the common causes of 3D reconstruction and tracking failure, as these processes require continuous tissue visibility.

In most of the recent vision-based deformation recovery methods, tissue deformation is usually represented by the displacement field~\cite{giannarou_vision-based_2016, chen_trans-window_2024, wang_neural_2022}, which merely has 3-DoF. However, surface rotation and inner-surface stretching are not modeled, which are up to 12-DoF and are necessary to represent 3D surface deformation from the perspective of continuum mechanics. Therefore, we reformulate the commonly used definition of tissue deformation in a vision-based approach by adding the additional parameters to represent pointwise rotation and stretching. We also propose a camera-centric setting to get rid of the dependence on camera pose. Based on all these prerequisites, we proposed an online method to concurrently optimize tissue geometry and deformation leveraging the tissue-invariant vision-biomechanics consistency constraint.

\subsection{Related work}
{\bf{3D reconstruction:}} 3D scene reconstruction is the basis of tissue deformation recovery, which provides tissue 3D structure as a depth map, point cloud, or mesh. Typical sensors for vision-based 3D reconstruction are either monocular or stereo (binocular) cameras. Monocular image-based 3D reconstruction in a single frame is an ill-posed problem, as a single 2D frame loses the 3D information. Only relative depth can be estimated~\cite{budd_transferring_2024}. Scaling factors need to be estimated to convert the relative depth to the absolute one~\cite{chavarrias_solano_multi-task_2025, wei_enhanced_2024}, which remains challenging and the accuracy of which is typically lower than the stereo-based approach. Stereo image-based 3D reconstruction is typically formulated as a stereo matching problem. Given a single pair of the left and right frames, it is theoretically sufficient to recover a 3D structure~\cite{xia_robust_2022, psychogyios_msdesis_2022, bardozzo_stasis-net_2022}. However, due to occlusion, smoking, lowly textured surfaces, duplicated textures, and specular reflection problems, mismatches always occur, leading to erroneous reconstructed results.

{\bf{Inter-frame alignment:}} Inter-frame alignment is necessary for continuous tissue deformation recovery. Tissue structures reconstructed from different frames may lie in different camera coordinates due to camera motion. Camera pose is used to represent camera orientation and position under the reference coordinate and can be used to transform frame data into the same coordinate. Camera pose estimation in robotic surgery can be realized by leveraging robot kinematics~\cite{li_super_2020, yoshimura_mbapose_2021}. However, kinematics data typically suffer from the accumulative error from the hand-eye calibration chain. Image-based camera pose estimation is another solution~\cite{luo_monocular_2023, hayoz_learning_2023}. However, camera pose estimation in a deformable scene is an ill-posed problem, as it is hardly possible to decouple camera and scene motion without identifying reliable rigid landmarks. Moreover, camera pose has only 6 DoF and is insufficient to describe the infinite DoF non-rigid deformation. Therefore, even if the camera pose can be perfectly estimated, it barely helps in tissue deformation recovery. To represent the infinite-DoF deformation, the displacement field is commonly used, which can describe the pointwise motion. Currently, the displacement field can be represented either implicitly or explicitly. An implicit displacement field is implemented using the neural field and rendering-based methods~\cite{wang_neural_2022, zha_endosurf_2023, yang_deform3dgs_2024, guo_free-surgs_2024, zhu_endogs_2025}. However, implicit displacement field is typically trained in an offline approach, and there is currently no evidence that it can achieve high tracking accuracy. Explicit displacement field is also named as scene flow. It can be estimated leveraging the 3D reconstruction and 2D tracking information~\cite{schmidt_tracking_2024, stoyanov_stereoscopic_2012, giannarou_vision-based_2016, chen_occlusion-robust_2023, chen_trans-window_2024, hayoz_online_2024}. Scene flow can easily represent the inter-frame motion of data points and thus is a good candidate for inter-frame alignment. However, vision-derived scene flow inherits the error of 3D reconstruction and 2D tracking and is typically rough and noisy. Deformable registration is another widely used approach for inter-frame alignment~\cite{wang_video-based_2024, yang_boundary_2025}, while these methods typically focus on the registration between preoperative and intraoperative frames, and thus, are not within the scope of our discussion.

{\bf{Simultaneous reconstruction and alignment:}} The tasks of 3D reconstruction and inter-frame alignment are often treated as one task. Under such problem settings, the most famous methods are the simultaneous localization and mapping (SLAM) and structure from motion (SfM) families~\cite{gomez-rodriguez_sd-defslam_2021, lamarca_defslam_2021, rodriguez_nr-slam_2024, ozyoruk_endoslam_2021, chen_slam-based_2018, widya_3d_2019, shao_self-supervised_2022, recasens_endo-depth-and-motion_2021, shan_enerf-slam_2024}. These methods concurrently optimize camera pose and reconstruct 3D scenes and typically have long-term stability. Although these endoscopic SLAM and SfM methods are adapted to deformable scenes by adding some flexibility in modeling structure deformation, they still cannot handle large deformation and are not robust to tool-induced occlusion. More importantly, these methods focus more on reconstructing panoramic scenes while paying little attention to continuous point tracking in the time domain.

{\bf{Biomechanics-driven deformation estimation:}} Biomechanical properties, such as stiffness and elasticity, are used in modeling the tissue dynamic behavior. Once the tissue-specific biomechanical property is available, a deformation model can be established to estimate and predict tissue deformation under tool-tissue interaction. Such technique is commonly used in surgical simulation~\cite{scheikl_sim--real_2023, zhu_real-time_2022} and also appears to be used in processing the real deformable object videos~\cite{haouchine_vision-based_2018, haouchine_monocular_2015}. However, tissue-specific and patient-specific tissue biomechanical properties are usually unavailable in real surgery, limiting the use of the biomechanics-driven method. Additionally, estimating tissue property from endoscopic images becomes possible by leveraging vision-derived deformation~\cite{liang_real--sim_2024, xu_identification_2022}; however, this turns out to be a chicken-and-egg problem.

\subsection{Contributions}
The proposed method overcomes the limitations of previous vision-based deformation recovery methods. Key contributions are as follows.
\begin{enumerate}
\item{{\bf{Deformation representation:}} It is the first time that deformation analysis with displacement, local surface rotation, and stretching can be conducted in intraoperative endoscopic imaging, with the novel 9-DoF deformation descriptor.}
\item{{\bf{Geometry representation:}} A novel 9-DoF geometric descriptor is proposed that represents both 3D position and differential geometry features related to local deformation.}
\item{{\bf{Camera-pose-free:}} The proposed method is formulated under a novel camera-centric setting and achieves truly camera-pose-free. The method neither requires the camera pose as input, estimates it, nor uses it in reconstruction.}
\item{{\bf{Tissue-invariant:}} The method does not rely on tissue-specific biomechanical priors.}
\item{{\bf{Online and occlusion robust:}} The method works in an online manner and does not involve offline optimization. Historical and the latest data are maintained under the proposed canonical mechanism. Such online behavior enables the use of historical data to estimate the currently unobservable data, making the method robust to occlusion.}
\end{enumerate}

\section{Preliminaries}

\subsection{Tissue surface deformation representation}\label{sec_definition}

Previously, tissue deformation is often represented by various displacement fields, such as the neural field~\cite{pumarola_d-nerf_2021, wang_neural_2022, zha_endosurf_2023}, pointwise displacements~\cite{chen_occlusion-robust_2023, stoyanov_stereoscopic_2012, giannarou_vision-based_2016}, or displacement maps~\cite{chen_trans-window_2024}. These displacement fields describe the 3D displacements of the tissue surface points between frames or relative to a reference frame. However, they cannot directly tell the deformation within a given region on the tissue surface, since the 3D displacement $(\Delta x,\Delta y,\Delta z)\in\mathbb{R}^3$, as a 3-DoF descriptor, is not enough to describe 3D deformation.

\begin{figure}[!t]%
\centering
\includegraphics[width=0.49\textwidth]{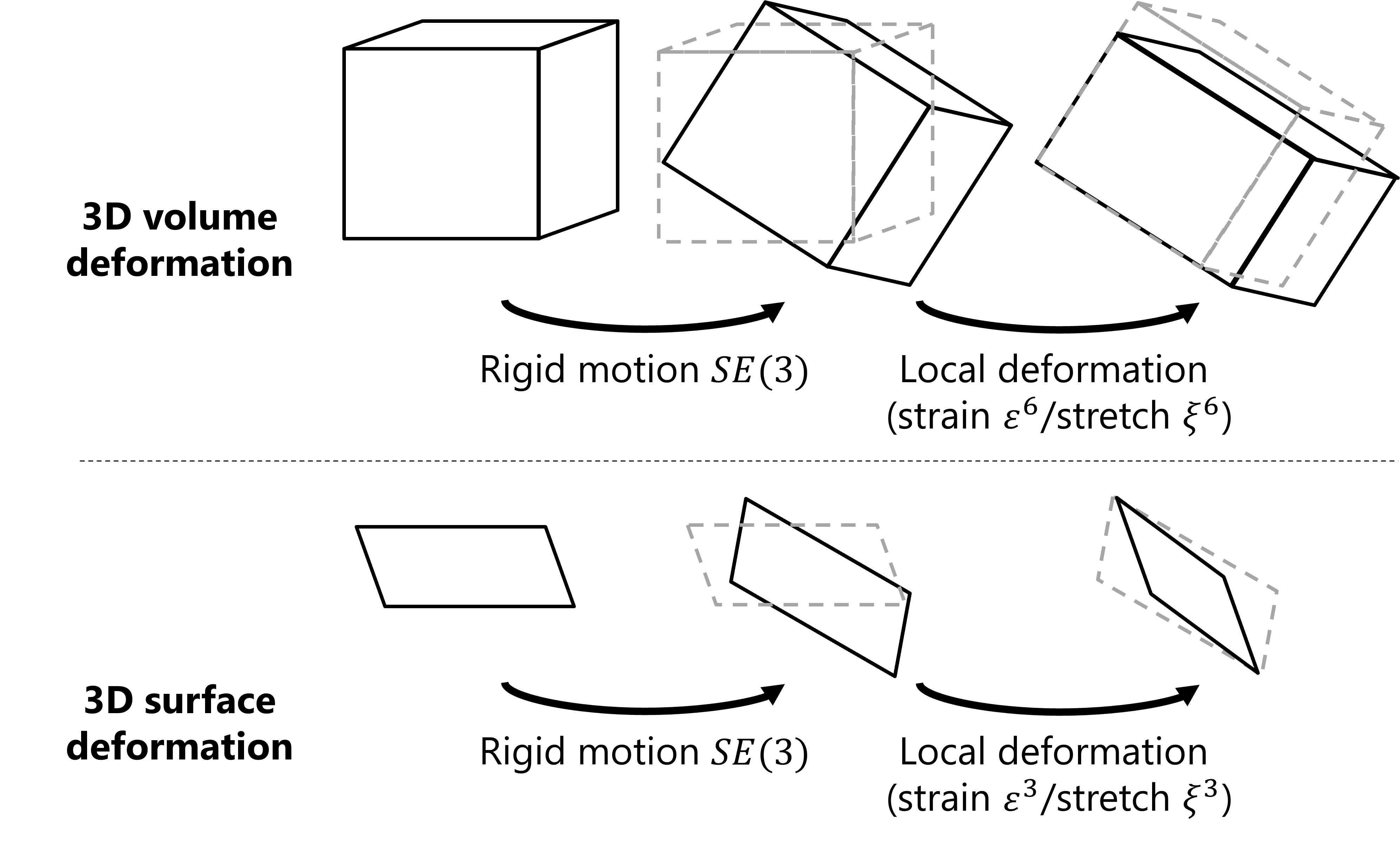}
\caption{The representation of deformation in the 3D space.} \label{fig_deformation}
\end{figure}

In continuum mechanics~\cite{spencer_2004_continuum}, we need a 12-DoF descriptor to represent the 3D deformation for a volume, as shown in Fig.~\ref{fig_deformation}, where 6-DoF are used to represent the rigid motion and the other 6-DoF are for the local deformation. The 12-DoF 3D deformation descriptor can be described as $\bm{d}\in \{SE(3),{\varepsilon}^6\}$, where $SE(3)$ is the Lie group representing the homogeneous transformation of the rigid motion, and ${\varepsilon}^6$ represents the strain, including three normal strains and three shear strains. Since the polar decomposition in continuum mechanics can directly get the stretch rather than the strain, for convenience, we use stretch instead of strain in the following text, such that the 3D deformation is as $\bm{d}\in \{SE(3),{\xi}^6\}$, where ${\xi}^6$ represents the stretch. One can transform stretch to strain without effort. For a 3D surface, as shown in Fig.~\ref{fig_deformation}, the representation of 3D deformation can be simplified to $\bm{d}\in \{SE(3),{\xi}^3\}$.

The description of local deformation is rotation-variant; for convenience, we define $\mathbb{D}=(r^3,\xi^3)$ as a local deformation descriptor, where $r^3=\mathbb{R}^3$ describes the 3-DoF rotation and $\xi^3$ is the inner-surface stretch. $r^3$ represents three independent rotation Euler angles that are interpreted in order as a rotation around the z-axis, a rotation around the y-axis, and a rotation around the x-axis, which can be converted to rotational transformation $SO(3)$. Finally, the 3D surface deformation is as $\bm{d}\in \{\mathbb{F},\mathbb{D}\}$, where the $\mathbb{F}=\mathbb{R}^3$ stands for 3D displacement. We claim this as a sufficient representation for 3D surface deformation.

\subsection{Camera-pose-free scene reconstruction}
Camera pose represents the orientation and position of a camera with respect to a reference coordinate. The term ``camera-pose-free'' in previous methods~\cite{guo_free-surgs_2024, recasens_endo-depth-and-motion_2021, gomez-rodriguez_sd-defslam_2021} often refers to the unnecessity of the camera pose as input, while the camera pose is still crucial and indispensable for scene reconstruction and is estimated inside their algorithms. Different from the previous one, the meanings of ``camera-pose-free'' in this article are twofold: 1. the camera pose is unnecessary as input; 2. the camera pose is totally not considered in the problem formulation nor used in reconstruction.

The motivation of totally replacing camera pose is twofold. First is the remaining challenge in accurately acquiring camera pose in surgical intervention~\cite{li_super_2020, yoshimura_mbapose_2021, hayoz_learning_2023}. Second is that even an error-free camera pose\footnote{In the case where the camera is fixed, we can say there is no error in camera pose.} barely helps in tissue deformation recovery without pointwise tracking or registration. Thus, pointwise tracking can be regarded as a higher level of data that enables inter-frame alignment as compared to camera pose.

\begin{figure}[!t]%
\centering
\includegraphics[width=0.45\textwidth]{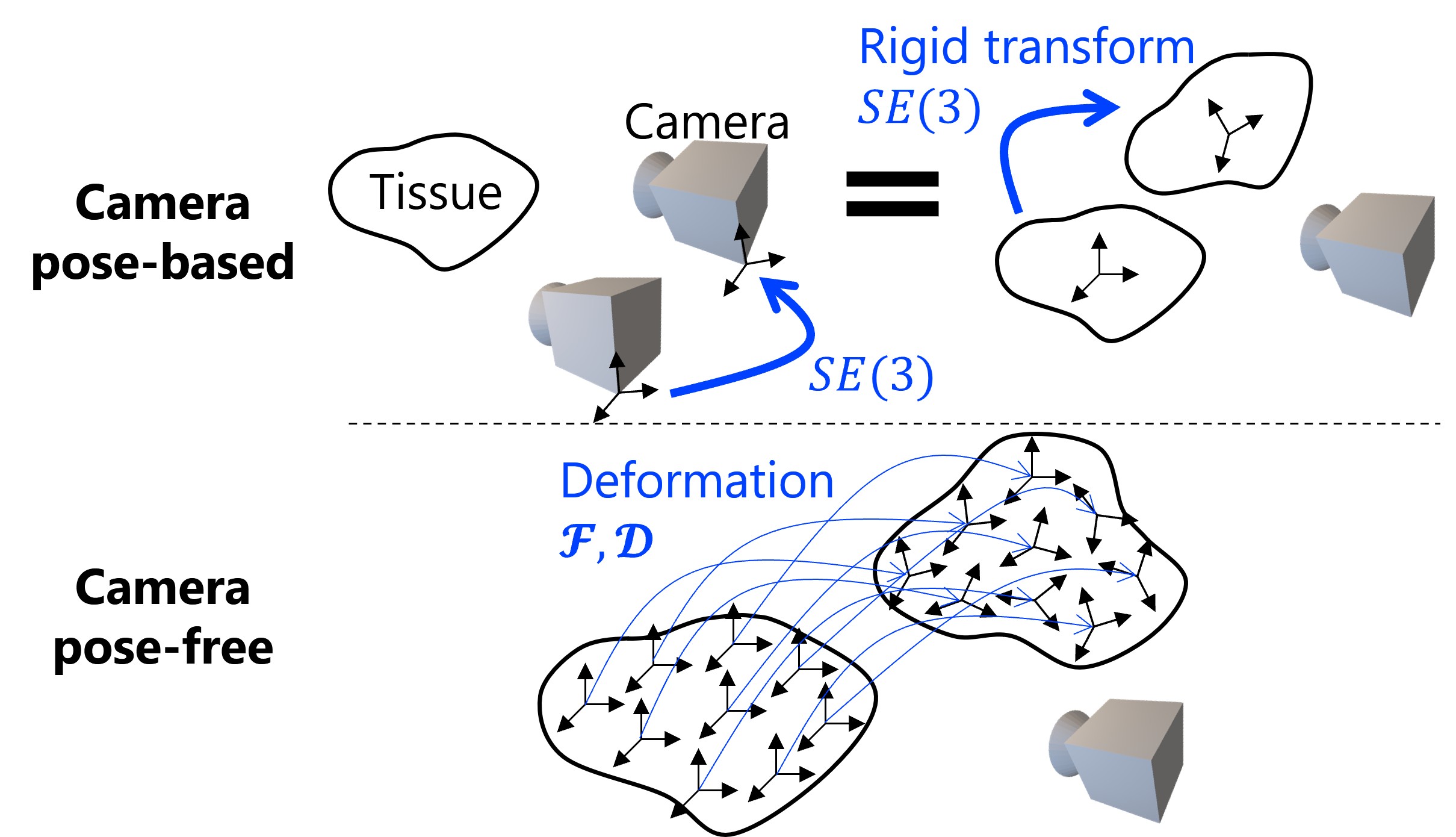}
\caption{Illustration of the proposed camera-pose-free approach in the inter-frame alignment, as compared to camera-pose-based ones.}\label{fig_camera_pose_free}
\end{figure}

Therefore, in this article, we do not attempt to estimate the camera pose or decouple the motion of the scene and the camera. Instead, eliminating the camera pose gives us a novel camera-centric problem setting. In the frame-by-frame deformation recovery process, the geometry and deformation are always represented within the latest camera coordinates. All subsequent recovery of the deformable structure are incremental based on the first frame coordinates, while the inter-frame alignment is achieved via dense pointwise modeling of the inter-frame deformation. As illustrated in Fig.~\ref{fig_camera_pose_free}, the camera-pose-based approach leverages 6-DoF to represent rigid scene motion, while this study employs 9$N$-DoF to represent deformable scene motion, where $N$ is the number of tissue surface points. This fundamental difference enables a truly camera-pose-free way to efficiently model a deformable scene.

\subsection{Problem setting}

{\bf{Prerequisites and assumptions:}}
\begin{enumerate}
  \item The proposed method relies on two measurements: the 3D scene structure and 3D tracking, typically the 3D point cloud and the 3D scene flow map. These measurements are always noisy. Noises may come from specular reflection, duplicated texture, low-textured regions, self-occlusion, tool-induced occlusion, camera motion, and tissue sudden breaking.
  \item The mask generated from instrument segmentation is accurate or can at least cover the whole region of the instrument in the image.
  \item The method is online. Only the latest and the second-latest frame data are involved in optimization.
  \item The camera can be either fixed or moving, while the camera pose is always unnecessary and will not be estimated.
  \item Tissue surfaces may temporarily or permanently appear or disappear due to camera motion and surgical manipulation and can also be cut through in the case of dissection.
  \item Vision-biomechanics consistency assumption: inter-frame deformation is limited, and the deformation between neighboring local surfaces is similar.
\end{enumerate}

{\bf{Objectives:}} Given the measured 3D scene structure and 2D tracking, we want to estimate the geometry and the deformation $\bm{d}\in \{\mathbb{F},\mathbb{D}\}$ of the tissue surface. Tissue geometry and deformation can be estimated even if the region is currently occluded by surgical instruments or is located outside the field of view.

\begin{figure*}[!t]%
\centering
\includegraphics[width=0.68\textwidth]{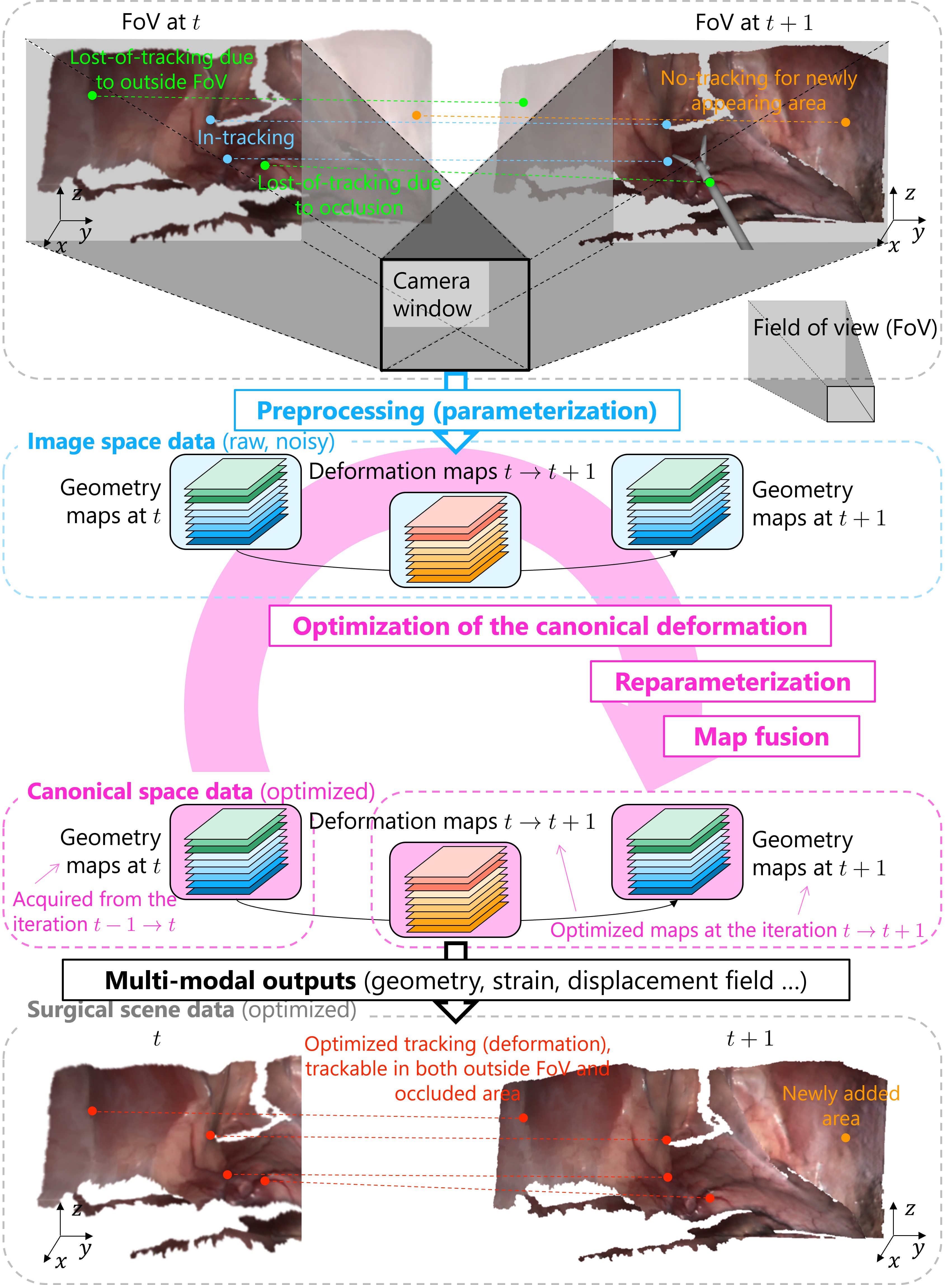}
\caption{Schematic of the proposed method, showing how the method works within one iteration.} \label{fig_overview}
\end{figure*}

\section{Methods}\label{Methods}

\subsection{Overview}

We present a vision-based tissue deformation recovery method with robustness to occlusion and camera motion under a camera-centric problem setting. Tissue geometry and deformation used in the method are mapped in either the canonical space or the image space, as shown in Fig.~\ref{fig_overview}. The method assumes that the tissue deformation is partially observable (trackable) and optimizes the geometry and deformation with tissue connectivity and vision-biomechanics consistency. Such optimization is performed in an online approach: when a new frame comes, the frame map together with the last recovered map gets involved in the optimization. The newly appearing area in the new frame will then be fused into the optimized canonical map. Such design brings us the two major benefits: 1. the method can handle camera motion while not being influenced by camera pose estimation error; 2. the method is robust to occlusion and can always reveal the latest observation, such as the instant change in tissue texture and shape. The optimized geometry and deformation maps can be converted into multi-modal data including point cloud, strain distribution, and displacement field for visualization and further analysis.

\begin{figure}[!t]%
\centering
\includegraphics[width=0.45\textwidth]{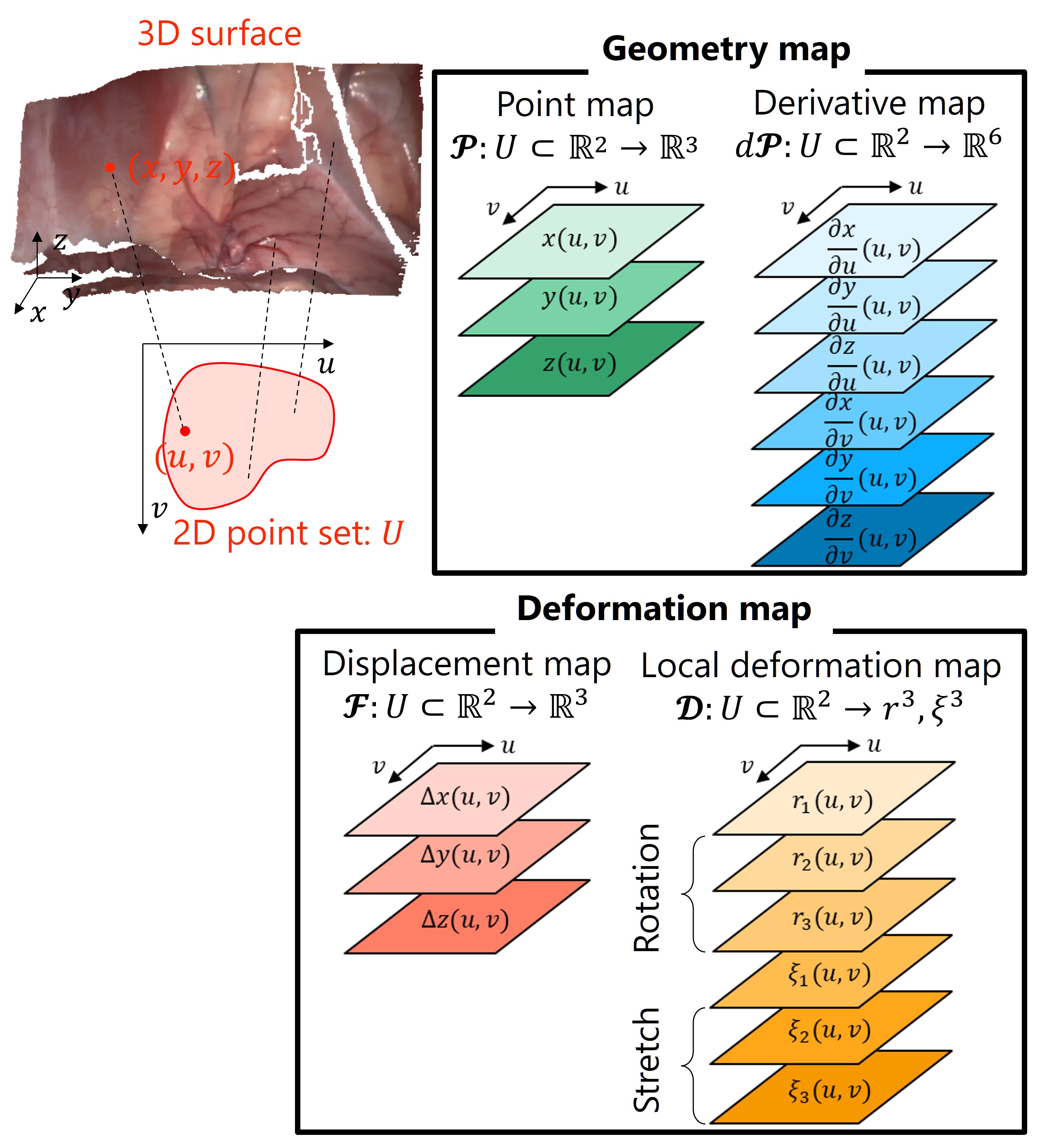}
\caption{The parameterization of a 3D surface and the corresponding geometry and deformation map.} \label{fig_parameterization}
\end{figure}

\subsection{Parameterization of 3D surface geometry and deformation}\label{sec_parameterization}
In this article, a parameterization is a map to describe 3D geometry and deformation using 2D parameters. Parameterization helps to reduce the dimensional complexity required for describing the 3D geometry and deformation and to establish connectivity relationships among data points, which is crucial in optimization.

\begin{figure}[!t]%
\centering
\includegraphics[width=0.48\textwidth]{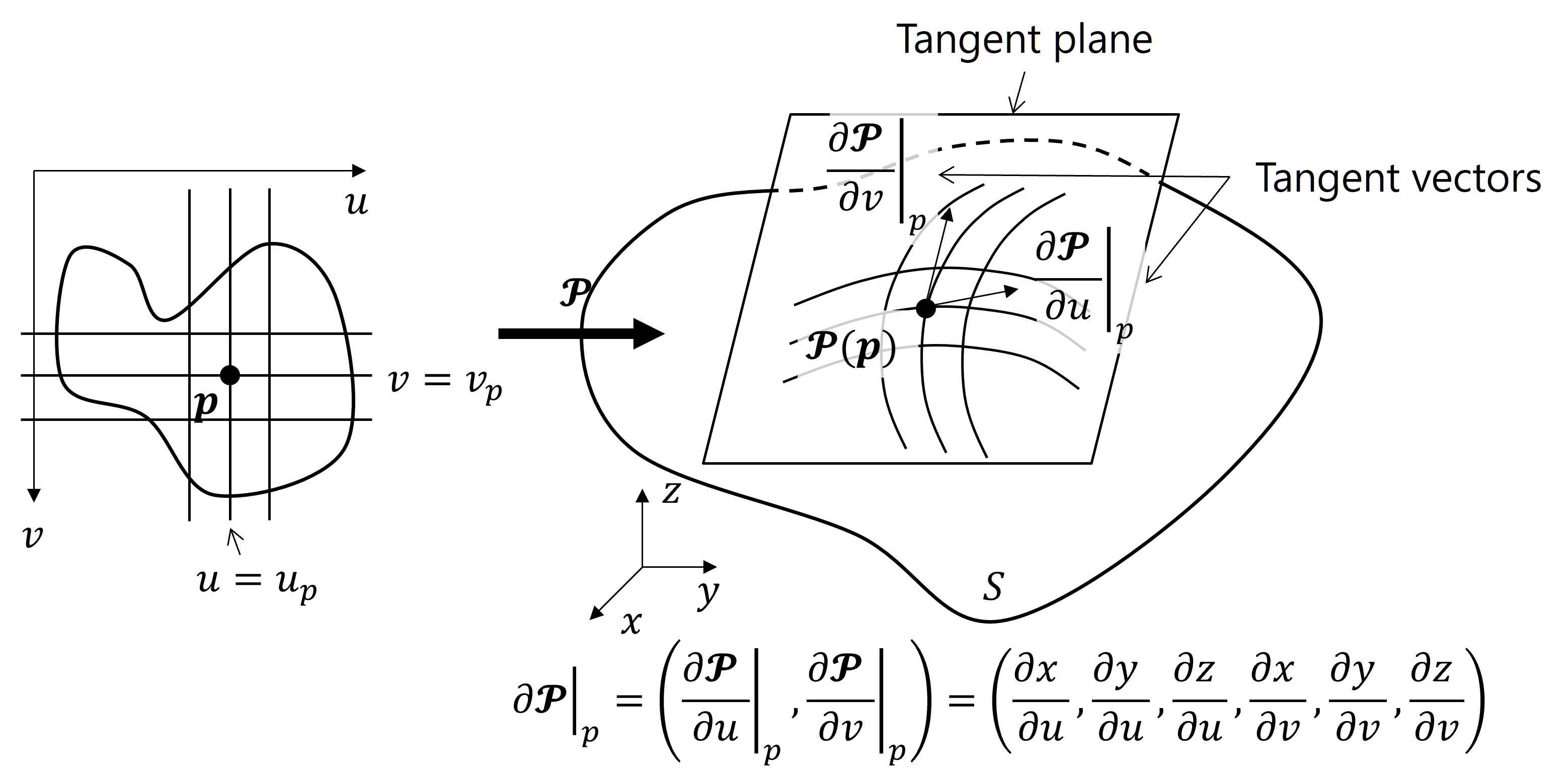}
\caption{The geometric meaning of the derivative of a 3D point map in the sense of differential geometry.} \label{fig_differential_geometry}
\end{figure}

Given a 3D surface geometry $S=\{(x,y,z)|x,y,z\in\mathbb{R}\}$, such as a 3D point cloud or a mesh, a commonly used method for parameterization is to project the 3D points/vertices onto a 2D plane, resulting in a set of 2D points $U=\{(u,v)|u,v\in\mathbb{R}\}$, which creates a 2D-3D map $\bm{\mathcal{P}}:U\to S$, as shown in Fig.~\ref{fig_parameterization}. If the 3D geometry is provided as a depth/disparity image~\cite{lipson_raft-stereo_2021, xia_robust_2022}, or a 3D point map~\cite{chen_trans-window_2024}, then naturally these can be regarded as a parameterization of 3D geometry. To avoid sharp edges and discontinuity in the structure, we parameterize only the subset of the 3D surface, such that $\bm{\mathcal{P}}$ is a homeomorphism and differentiable. Accordingly, we can define the derivative of $\bm{\mathcal{P}}$ as: 
\begin{equation}\label{eq_dM}
d\bm{\mathcal{P}} (u,v)=\left(\frac{\partial x}{\partial u}\ \frac{\partial y}{\partial u}\ \frac{\partial z}{\partial u}\ \frac{\partial x}{\partial v}\ \frac{\partial y}{\partial v}\ \frac{\partial z}{\partial v}\right)
\end{equation}
which form the tangent plane of the surface $S$ at the point $\bm{p}=\bm{\mathcal{P}}(u,v)$, representing local surface orientation and inclination~\cite{manfredo_differential_2016}, as shown in Fig.~\ref{fig_differential_geometry}. In this article, both the 3D point map $\bm{\mathcal{P}}$ and its derivative map $d\bm{\mathcal{P}}$ are used to represent surface geometry, so called geometry map, as shown in Fig.~\ref{fig_parameterization}.

The deformation in this article is defined as $\bm{d}\in \{\mathbb{F},\mathbb{D}\}$, which contains the 3D displacement $\mathbb{F}$ and the local deformation $\mathbb{D}$. 3D displacement can be directly obtained from the measurement. Let's first ignore occlusion and assume each data point has a 3D displacement, such that we have a displacement map $\bm{\mathcal{F}}:U\to\mathbb{F}$. After displacement, a map to describe the new point locations is $\bm{\mathcal{P}}'=\bm{\mathcal{P}}+\bm{\mathcal{F}}$. Consequently, the derivative of $\bm{\mathcal{P}}'$ is $d\bm{\mathcal{P}}'$. We can then calculate the local deformation $\bm{d}\in\mathbb{D}$ containing surface rotation and inner-surface stretching. Finally, the parameterized tissue local deformation is as $\bm{\mathcal{D}}:U\to \mathbb{D}$. The 3D displacement map $\bm{\mathcal{F}}$ and the local deformation map $\bm{\mathcal{D}}$ together are named deformation map, as shown in Fig,~\ref{fig_parameterization}.

\begin{figure}[!t]%
\centering
\includegraphics[width=0.45\textwidth]{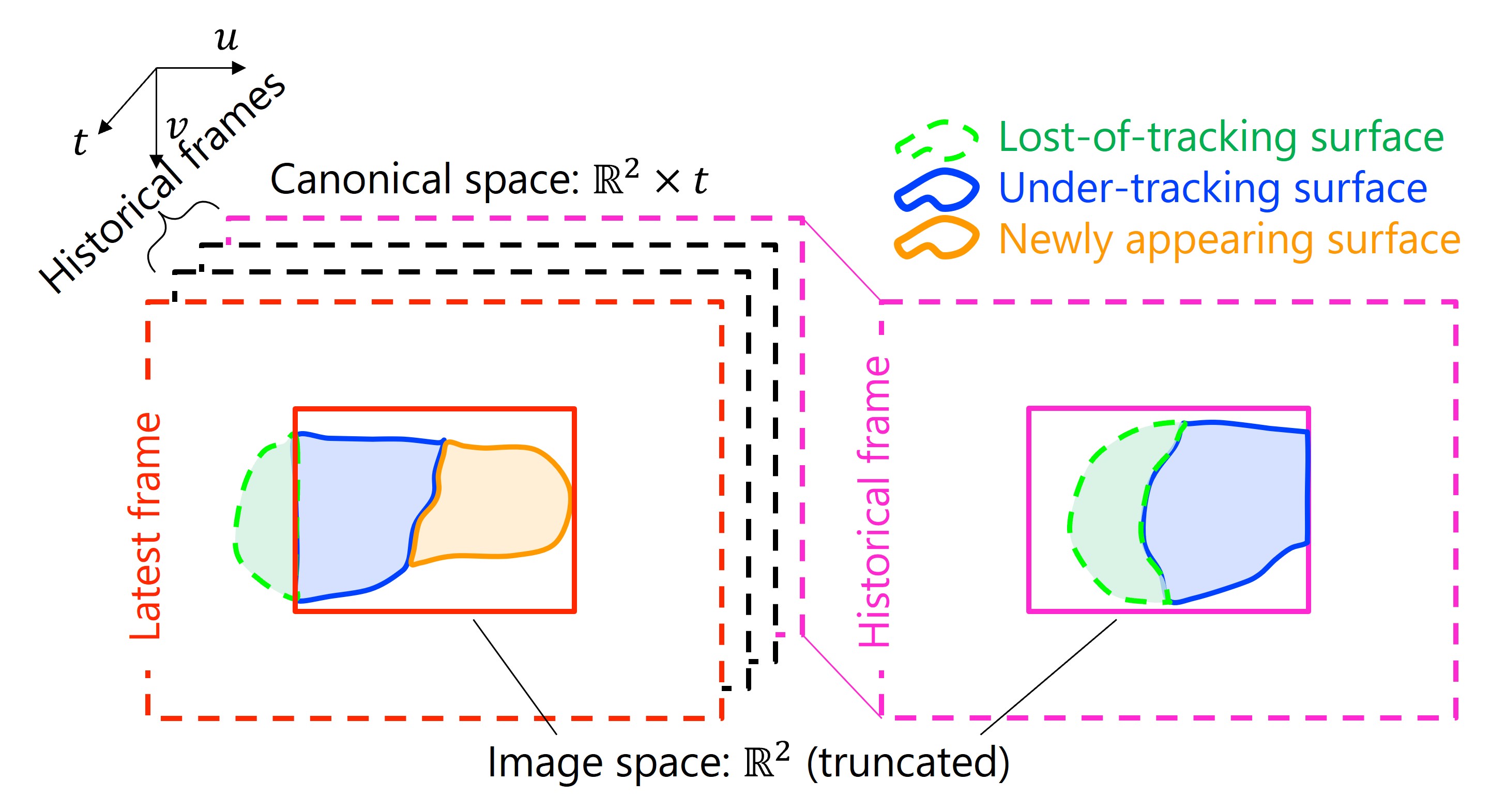}
\caption{Illustration of the canonical space and the image space. The canonical space is a borderless 2D space with time domain and keeps both the historical and the latest data, while the image space is a truncated 2D space, such that information in the image space may get lost.} \label{fig_canonical}
\end{figure}

\subsection{Image space and canonical space}

Maps from the 2D space to higher dimensional space have been established via parameterization. Intuitively, the 2D space will be the image space, which is a truncated 2D space due to the physical limitation of the field of view (FoV). However, image space throws away data once the object moves outside the FoV. To address this problem, a canonical space is proposed for maintaining both the historical and the latest observed data. The canonical space is a borderless 2D space with a time dimension where all maps of the finalized reconstruction are defined. As a camera-centric setting, the per-frame image space is always aligned at the center of the canonical space, as shown in Fig.~\ref{fig_canonical}. The data in the image space always reveal the latest observation, while the data in the canonical keeps the latest optimized as well as the historical data.

\subsection{Notation and conventions}~\label{section_notation}

For clarification, this section explains the naming rules of the variables used in the algorithm. Most of the variables are maps from the parameter space, including the image space and the canonical space, to the higher dimensional space representing various modeled quantities, including 3D point positions, surface derivatives, 3D displacement (scene flow), 2D displacement (optical flow), local deformation, and texture. As shown in Table~\ref{tab_notation}, depending on whether these maps are defined in the image or the canonical parameter spaces, their subscripts are ``I'' or ``C'', respectively. To distinguish maps at different frames, additional subscript representing time is added. For example, $\bm{\mathcal{P}}_I|_{t+1}$ represents a point map defined in the image space at time $t+1$, while $\bm{\mathcal{F}}_C|_{t}$ represents a displacement map defined in the canonical space at time $t$. Note that $t$ refers to a discretized measurement frame index rather than physical time.

\begin{table*}[!t]
   \caption{Notation of maps}
   \label{tab_notation}
   \scriptsize
   \begin{center}
   \begin{tabular}{lcll}
   \toprule
   \multicolumn{2}{c}{\diagbox[]{n-D space}{2D parameter space}} & \makecell[c]{Image \\ $U_I=\{(u,v)|u,v\in\mathbb{R}\}$} & \makecell[c]{Canonical \\ $U_C=\{(u,v)|u,v\in\mathbb{R}\}$} \\
   \midrule
   3D point & $\mathbb{R}^3$ & ~~~~~$\bm{\mathcal{P}}_I:U_I\to\mathbb{R}^3$ & ~~~~~$\bm{\mathcal{P}}_C:U_C\to\mathbb{R}^3$ \\
   \midrule
   Surface derivative (local geometry) & $\mathbb{R}^6$ & ~~~~$d\bm{\mathcal{P}}_I:U_I\to\mathbb{R}^6$ & ~~~~$d\bm{\mathcal{P}}_C:U_C\to\mathbb{R}^6$ \\
   \midrule
   3D displacement (scene flow) & $\mathbb{F}=\mathbb{R}^3$ & ~~~~~$\bm{\mathcal{F}}_I:U_I\to\mathbb{F}$ & ~~~~~$\bm{\mathcal{F}}_C:U_C\to\mathbb{F}$ \\
   \midrule
   \makecell[l]{Local deformation \\ (rotation and stretching)} & $\mathbb{D}=(r^3,\xi^3)$ & ~~~~~$\bm{\mathcal{D}}_I:U_I\to\mathbb{D}$ & ~~~~~$\bm{\mathcal{D}}_C:U_C\to\mathbb{D}$ \\
   \midrule
   2D displacement (optical flow) & $\mathbb{R}^2$ & ~~~~~$\bm{\mathcal{O}}_I:U_I\to\mathbb{R}^2$ & ~~~~~$\bm{\mathcal{O}}_C:U_C\to\mathbb{R}^2$ \\
   \midrule
   Texture & $RGB$ & ~~~~~~$\bm{\mathcal{C}}_I:U_I\to RGB$ & ~~~~~~$\bm{\mathcal{C}}_C:U_C\to RGB$ \\
   \bottomrule
   \end{tabular}
   \end{center}
\end{table*}

\subsection{Algorithm overview}

The proposed modeling algorithm works in an online approach. In each iteration, inter-frame data get involved in the calculation. Let's say the current frame is at time $t+1$ and the last frame is at time $t$. The algorithm in each iteration involves three three steps: 1). optimization of the canonical deformation; 2). reparameterization; and 3). map fusion. The algorithm is illustrated in Fig.~\ref{fig_algorithm} (see appendix~\ref{appendix_pseudo} for the pseudocode).

{\bf{Inputs:}} The proposed method takes two types of measurements as inputs: 3D structure and tracking. The measurement of 3D structure can be acquired via depth estimation, such as stereo matching~\cite{lipson_raft-stereo_2021}, while the measurement of 3D tracking can be acquired via scene flow or a two-step optical flow approach~\cite{chen_occlusion-robust_2023, stoyanov_stereoscopic_2012}. The measurements are noisy and are parameterized into maps under the bounded image space. These include the 3D point maps ($\bm{\mathcal{P}}_I|_{t}$, $\bm{\mathcal{P}}_I|_{t+1}$), derivative maps ($d\bm{\mathcal{P}}_I|_{t}$, $d\bm{\mathcal{P}}_I|_{t+1}$), and deformation maps between frame $t$ and $t+1$ ($\bm{\mathcal{F}}_I|_{t}$ and $\bm{\mathcal{D}}_I|_{t}$). Detailed preprocessing algorithm is introduced in appendix~\ref{appendix_pseudo}. All these measurement maps in the image space together with the previously modeled canonical geometry maps $\bm{\mathcal{P}}_C|_{t}$ and $d\bm{\mathcal{P}}_C|_{t}$ are used as the inputs of the current iteration ($t+1$). If the frame at $t$ is the initial frame, the canonical geometry maps are initialized to be the same as $\bm{\mathcal{P}}_I|_{t}$ and $d\bm{\mathcal{P}}_I|_{t}$. Note that parts of these maps not belonging to the tissue but to the surgical instrument are marked and removed using surgical instrument segmentation.

{\bf{Outputs:}} The final outputs of the algorithm are the canonical deformation maps\footnote{These are ``forward deformation''. They describe how the data points at $t$ deform.} $\bm{\mathcal{F}}_C|_{t}$ and $\bm{\mathcal{D}}_C|_{t}$ and the canonical geometry maps $\bm{\mathcal{P}}_C|_{t+1}$ and $d\bm{\mathcal{P}}_C|_{t+1}$.

\begin{figure*}[!t]%
\centering
\includegraphics[width=0.7\textwidth]{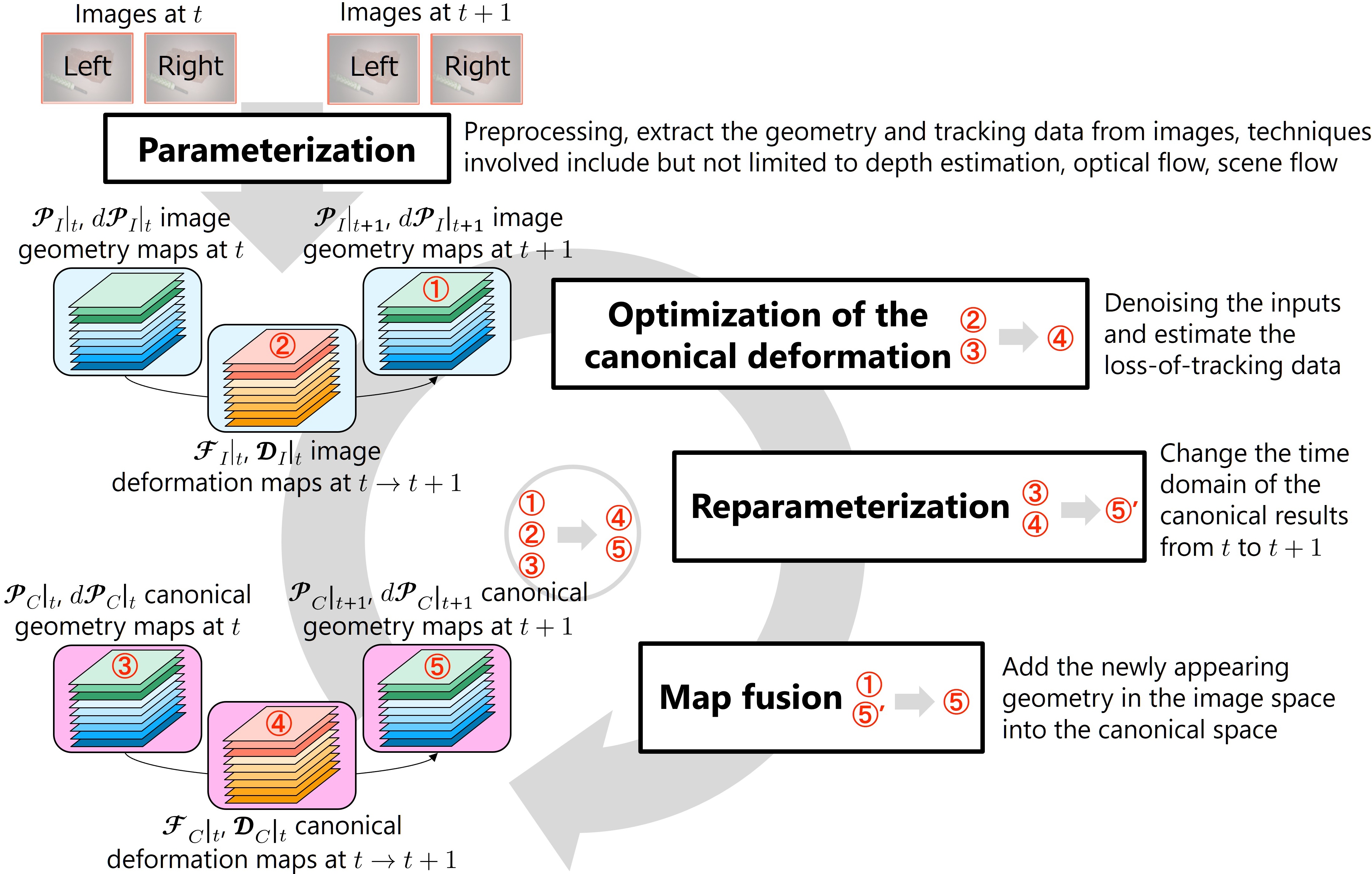}
\caption{Illustration of the deformation recovery algorithm per iteration.} \label{fig_algorithm}
\end{figure*}

\subsection{Optimization of the canonical deformation}\label{sec_optimization}

The desired output deformation maps $\bm{\mathcal{F}}_C|_{t}$ and $\bm{\mathcal{D}}_C|_{t}$ are defined on the canonical parameter set $U_C|_t$. Parts of the canonical deformation maps can be initialized using the input deformation maps $\bm{\mathcal{F}}_I|_{t}$ and $\bm{\mathcal{D}}_I|_{t}$ defined on the image parameter set $U_I|_t$. The initializable deformation is also called the observable deformation. However, since $U_C|_t\neq U_I|_t$ due to tracking loss, $\bm{\mathcal{F}}_C|_{t}$ and $\bm{\mathcal{D}}_C|_{t}$ is not fully defined in $U_C|_t$. Therefore, to acquire a fully defined canonical deformation map, we design an optimization method leveraging the observable deformation. In the following sections, we first introduce tissue-invariant vision-biomechanics consistency and then show how it helps outlier detection and deformation optimization.

\subsubsection{Tissue-invariant vision-biomechanics consistency}\label{sec_consistency} 
Vision-derived measurement of deformation $(\bm{\mathcal{F}}_I|_{t},\bm{\mathcal{D}}_I|_{t})$ is typically noisy, sensitive to occlusion, and limited in the field of view. The noises in visual measurement may come from various sources, such as specular highlight, occlusion, and camera motion~\cite{chen_trans-window_2024, zhang_specular_2023}. It is difficult to quantify all the error sources and eliminate the corresponding noises specifically. This urges us to propose a more universal denoising solution. Fortunately, what's for sure is that the tissue deformation we observed in surgical videos has to be consistent with the biomechanism. Although many biomechanical properties are tissue-specific~\cite{haouchine_vision-based_2018, scheikl_sim--real_2023}, making them hardly possible to be applied to universal intraoperative cases, we still manage to summarize some that are tissue-invariant. To ensure the vision-derived deformation is biomechanically reasonable, we introduce the following assumptions.

{\bf{Limitation of the inter-frame strain:}} Given the truth that a common frame-per-second (FPS) of the surgical video is higher than 30, we can assume that the deformation, or strain, between frames should be trivial. Empirically, the principal strain should not be smaller than $-0.1$ nor larger than $0.1$. This gives us a very rough yet efficient constraint to limit the inter-frame deformation. Local disturbance caused by various error sources typically results in a relatively large strain and thus can be detected. Please refer to appendix~\ref{appendix_strain} for more detailed justification of the inter-frame strain threshold.

{\bf{Similarity of neighboring deformation:}} Given two adjacent locations, $\bm{p}_1$ and $\bm{p}_2$, on the surface, we assume their corresponding local deformation, $\bm{d}_1$ and $\bm{d}_2$, are similar. In this article, the local deformation is represented as $\bm{d}\in\mathbb{D}$ and can be divided into rotation $\bm{r}\in r^3$ and stretch $\bm{\xi}\in\xi^3$. Therefore, we have $\bm{r}_1\approx\bm{r}_2$ and $\bm{\xi}_1\approx\bm{\xi}_2$. This assumption is based on the connectivity of surface geometry and the continuity of deformation and regards local tissue biomechanical properties to be the same.

\subsubsection{Outlier detection}
Outliers in the canonical deformation map result in inaccuracy and instability and thus need to be filtered out before moving on to the optimization. Each data point $(u,v)$ in the canonical parameter set $U_C|_{t}$ is either an inlier or an outlier. The unobservable data points $(u,v)\in U_C|_{t}\backslash U_I|_t$, including the occluded ones and the outside-field-of-view ones, are regarded as outliers. The tool-induced occlusion can be indicated by the mask derived from instrument segmentation~\cite{wang_neural_2022, psychogyios_msdesis_2022, wang_lacoste_2025}. Besides, there also exist outliers in the observable data points $(u,v)\in U_C|_{t}\cap U_I|_t$ that do not satisfy the vision-biomechanics consistency. For example, if the principal strain component in $\bm{\mathcal{D}}_I|_{t}(u,v),(u,v)\in U_C|_{t}\cap U_I|_t$ exceeds the inter-frame strain limitation as discussed in section~\ref{sec_consistency}, the data point is marked as an outlier. Finally, data points in $U_C|_{t}$ are separated into the inlier $U_{C,i}|_{t}$ and the outlier $U_{C,o}|_{t}$ sets. For the data points in the inlier set, their corresponding deformation should be similar to the observation, as:
\begin{equation}\label{eq_direct}
\begin{split}
\bm{\mathcal{F}}_{C}|_{t}(u,v)&\approx\bm{\mathcal{F}}_I|_{t}(u,v),\\
\bm{\mathcal{D}}_{C}|_{t}(u,v)&\approx\bm{\mathcal{D}}_I|_{t}(u,v), \text{for } (u,v)\in U_{C,i}|_{t}
\end{split}
\end{equation}

\subsubsection{Optimization of deformation}
The optimization of canonical deformation maps is performed for the local deformation map $\bm{\mathcal{D}}_C|_{t}$ first and then for the displacement map $\bm{\mathcal{F}}_C|_{t}$. This is because the regularization term in the optimization function for $\bm{\mathcal{F}}_C|_{t}$ involves the local geometry term that must be updated by $\bm{\mathcal{D}}_C|_{t}$.

{\bf{Local deformation:}} According to the assumptions of vision-biomechanics consistency in section~\ref{sec_consistency}, the local deformation of neighboring regions should be similar, as:
\begin{equation}\label{eq_local1}
    \bm{\mathcal{D}}_C|_{t}(\bm{p}_i)\approx \bm{\mathcal{D}}_C|_{t}(\bm{p}_j);i\not=j;\bm{p}_i,\bm{p}_j\in U_{C}|_{t}
\end{equation}
where $\bm{p}_i(u_i,v_i)$ and $\bm{p}_j(u_j,v_j)$ are adjacent 2D data points, such that we have $u_i=u_j\pm 1$ and $v_i=v_j\pm 1$ in a discrete form. To keep the consistency with the observation as in Eq.(\ref{eq_direct}), the optimization function for the local deformation map in the least square sense is:
\begin{equation}\label{eq_local2}
\begin{split}
\int_{\bm{p} \in U_{C,i}|_{t}}\left\| \bm{\mathcal{D}}_C|_{t}(\bm{p}) - \bm{\mathcal{D}}_I|_{t}(\bm{p}) \right\|^2 \\
+ \alpha\int_{\bm{p}_i,\bm{p}_j \in U_{C}|_{t}, i\not=j}\left\| \bm{\mathcal{D}}_C|_{t}(\bm{p}_i) - \bm{\mathcal{D}}_C|_{t}(\bm{p}_j) \right\|^2
\end{split}
\end{equation}
where Eq.(\ref{eq_local1}) is used as a regularization term ($\bm{p}_i$ and $\bm{p}_j$ are adjacent), and $\alpha$ is a scalar weight that may depends on the camera resolution and the physical scale of the tissue. To guarantee smoothness, we empirically fix $\alpha$ to $200$ in all our experiments. The estimated local deformation map $\widehat{\bm{\mathcal{D}}}_C|_{t}$ can be found that minimizes Eq.(\ref{eq_local2}), which is exactly a functional optimization problem and can be solved using general linear solvers or functional optimizer.

{\bf{Displacement:}} The displacement map is not directly optimized. Instead, we optimize the deformed 3D point map. The reason is because after optimizing the local deformation, we can update the local geometry and use it as a regularization term that preserves local smoothness and continuity. As mentioned in section~\ref{sec_parameterization}, we can deform the local geometry $d\bm{\mathcal{P}}_C|_{t}$ using $\widehat{\bm{\mathcal{D}}}_C|_{t}$ and acquire $ d\widehat{\bm{\mathcal{P}}}_C'|_{t}$, which is the estimated local geometry after deformation. If the estimated local deformation is correct, we will have:
\begin{equation}\label{eq_disp1}
d\bm{\mathcal{P}}_C'|_{t}(\bm{p}) \approx  d\widehat{\bm{\mathcal{P}}}_C'|_{t}(\bm{p});\bm{p}\in U_{C}|_{t}
\end{equation}
where $d\bm{\mathcal{P}}_C'|_{t}$ is the derivative of the deformed 3D point map. To keep the consistency with the observation as in Eq.(\ref{eq_direct}), the optimization function for the 3D point map in the least square sense is:
\begin{equation}\label{eq_disp2}
\begin{split}
\int_{\bm{p} \in U_{C,i}|_{t}}\left\| \bm{\mathcal{P}}'_C|_{t}(\bm{p}) - \bm{\mathcal{P}}_C|_{t}(\bm{p}) - \bm{\mathcal{F}}_I|_{t}(\bm{p}) \right\|^2 \\
+ \int_{\bm{p} \in U_{C}|_{t}}\left\| d\bm{\mathcal{P}}'_C|_{t}(\bm{p}) - d\widehat{\bm{\mathcal{P}}}'_C|_{t}(\bm{p}) \right\|^2
\end{split}
\end{equation}
where Eq.(\ref{eq_disp1}) is used as a regularization term. The optimized 3D point map $\widehat{\bm{\mathcal{P}}}'_C|_{t}$ is found that minimizes Eq.(\ref{eq_disp2}), which is again a functional optimization problem. After that, it is trivial to calculate the displacement map by $\widehat{\bm{\mathcal{F}}}_C|_{t} = \widehat{\bm{\mathcal{P}}}'_C|_{t} - \bm{\mathcal{P}}_C|_{t}$.

{\bf{Instrument pose-constrained optimization:}} If the pose of the surgical instrument is available, it can be used to constrain the optimization of geometry to improve the accuracy, especially in the contact area between the instrument tip and the tissue, as introduced in~\cite{chen_surgem_2024}. To leverage instrument pose constraints, we have to introduce the assumption that the target tissue is always under the instrument, that is: $z'>z_t$, where $z'$ is the depth component in $\bm{\mathcal{P}}'_C|_{t}(u,v)=(x',y',z')$ and $z_t$ is the tool depth of its bottom surface. Such a constraint can be applied to Eq.(\ref{eq_disp2}) to guide the optimization.

{\bf{Computational complexity:}} The proposed optimization is formulated as a least-squares problem over $N$ data points. The dominant computational cost arises from evaluating the residuals of the maps as in Eq.~\ref{eq_local2} and Eq.~\ref{eq_disp2}, alongside constructing and solving the resulting linear system. Because the maps $\bm{\mathcal{P}}'_C|_{t}(\bm{p})$ and $\bm{\mathcal{D}}_C|_{t}(\bm{p})$ are computed per-point, and the derivative $d\bm{\mathcal{P}}'_C|_{t}(\bm{p})$ relies strictly on local neighborhoods, the resulting system matrices are highly sparse and banded. Constructing this sparse system scales linearly with the number of data points, requiring $\mathcal{O}(N)$ operations. Subsequently, solving this sparse banded system utilizes standard sparse factorization, which has a theoretical complexity of $\mathcal{O}(N)$.

\subsection{Reparameterization}
After the optimization of canonical deformation, now we estimate the optimized canonical deformation maps ($\widehat{\bm{\mathcal{F}}}_C|_{t}$ and $\widehat{\bm{\mathcal{D}}}_C|_{t}$) and use them to deform the canonical geometry maps ($\bm{\mathcal{P}}_C|_{t}:U_C|_t\to \mathcal{S}_C|_{t}$ and $d\bm{\mathcal{P}}_C|_{t}:U_C|_t\to d\mathcal{S}_C|_{t}$), where $\mathcal{S}_C|_{t}=\{\mathbb{R}^3\}$ and $d\mathcal{S}_C|_{t}=\{\mathbb{R}^6\}$ stand for the geometry sets. The deformed canonical geometry maps are $\widehat{\bm{\mathcal{P}}}'_C|_{t}:U_C|_t\to \widehat{\mathcal{S}}'_C|_{t}$ and $d\widehat{\bm{\mathcal{P}}}'_C|_{t}:U_C|_t\to d\widehat{\mathcal{S}}'_C|_{t}$, where $\widehat{\mathcal{S}}'_C|_{t}=\{\mathbb{R}^3\}$ and $d\widehat{\mathcal{S}}'_C|_{t}=\{\mathbb{R}^6\}$ stand for the deformed geometry sets. However, all these maps are still defined in the canonical parameter set $U_{C}|_{t}$ rather than $U_{C}|_{t+1}$. In fact, $U_{C}|_{t+1}$ is still undefined so far. To move on to the latest frame at $t+1$, it is necessary to align the canonical space to the latest image space, that is, to generate parameter set $U_{C}|_{t+1}$ and define maps on it. Therefore, we parameterize the deformed geometry ($\widehat{\mathcal{S}}'_C|_{t}$ and $d\widehat{\mathcal{S}}'_C|_{t}$) following the same methods in section~\ref{sec_parameterization} and then acquire the 3D point map $\widehat{\bm{\mathcal{P}}}'_C|_{t+1}:\widehat{U}'_{C}|_{t+1}\to \widehat{\mathcal{S}}'_C|_{t+1}$ and its derivative map $d\widehat{\bm{\mathcal{P}}}'_C|_{t+1}:\widehat{U}'_{C}|_{t+1}\to d\widehat{\mathcal{S}}'_C|_{t+1}$. The texture map is bound to the point map and can be updated accordingly, as in appendix~\ref{appendix_pseudo}. Note that $\widehat{\mathcal{S}}'_C|_{t+1}$ is a subset of $\widehat{\mathcal{S}}'_C|_{t}$ as not all geometry can be parameterized in the projection-based approach\footnote{The geometry that is self-folding or located at the back of the camera cannot be parameterized.}.

\subsection{Map fusion}
Although we have deformed the canonical geometry map and aligned it to the latest frame, most of the data points in the map are inherited from the previous frames, which means that it cannot represent the newly appearing data. Fusion is to merge the newly appearing map defined in the latest image parameter set $U_I|_{t+1}$ to the one defined in the canonical parameter set $U_C|_{t+1}$.
A simple yet efficient strategy to merge these maps is if there is data in the canonical space, keep the canonical one; else, if there is data in the image space, keep the image one. This fusion strategy works for all maps in Table~\ref{tab_notation} except the texture map, for which we keep all data in the image space to preserve the latest observed tissue appearance. Please refer to appendix~\ref{appendix_pseudo} for more details.
Finally, both the updated historical data and the newly appearing data are merged together in the canonical space and will be involved in the optimization in the next iteration, as shown in Fig.~\ref{fig_algorithm}.

\subsection{Surface strain estimation}
In this section, we introduce how to leverage the recovered canonical maps to estimate tissue surface strain. We offer two ways to extract the strain data: the inter-frame strain and the accumulative strain.

{\bf{Inter-frame strain:}} The inter-frame canonical local deformation map $\bm{\mathcal{D}}_C|_t$ is the optimization target as discussed previously. For each data point $(u,v)\in U_C|_t$, a deformation descriptor $\bm{d}$ can be found as $\bm{d}=\bm{\mathcal{D}}_C|_t(u,v)\in\mathbb{D}$. $\bm{d}$ contains the stretch part, and it is trivial to turn stretch to strain. Therefore, the inter-frame strain can be obtained instantly from the recovered deformation map.

{\bf{Accumulative strain:}} More often, we want to continuously track surface points and observe their accumulative strain. In this study, the tracking of 3D surface points is realized by tracking the 2D data points in the canonical space. One can easily calculate the inter-frame optical flow map $\bm{\mathcal{O}}_C|_t$ given the inter-frame displacement map $\bm{\mathcal{F}}_C|_t$~\cite{chen_trans-window_2024}. Let's say we want to track the surface point from the $n$-th frame. To begin with, the canonical 3D point map at the $n$-th frame is $\bm{\mathcal{P}}_C|_n$. By iteratively applying the inter-frame optical flow map, we can track the 2D data points continuously, as:
\begin{equation}\label{eq_track}
    U_C|_{n\to n+m}=\bm{\mathcal{O}}_{C}|_{n+m-1} \circ ... \circ \bm{\mathcal{O}}_{C}|_{n}(U_C|_{n})
\end{equation}
Accordingly, the tracked 3D points are $\bm{\mathcal{P}}_C|_{n+m}(\bm{p}),\bm{p}\in U_C|_{n\to n+m}$. With the tracked point map, the accumulative deformation between the $n$-th frame and the ($n+m$)-th frame ($\bm{\mathcal{F}}_C|_{n\to n+m},\bm{\mathcal{D}}_C|_{n\to n+m}$) can be calculated according to section~\ref{sec_parameterization}. The stretch part in $\bm{\mathcal{D}}_C|_{n\to n+m}$ can be easily extracted and converted to strain, which is now the accumulative one.

\begin{table}[!t]
   \caption{Deformation recovery accuracy on the SURGEM dataset (unit in millimeter)}
   \label{tab_accuracy}
   \scriptsize
   \begin{center}
   \begin{tabular}{ccccc}
   \toprule
   \multirow{2}{*}{Methods} & \multicolumn{2}{c}{Non-occluded} & \multicolumn{2}{c}{Occluded} \\
   & RMSE ($\downarrow$) & MSD$\pm$std ($\downarrow$) & RMSE ($\downarrow$) & MSD$\pm$std ($\downarrow$) \\
   \midrule
   RAFT-Stereo~\cite{lipson_raft-stereo_2021} & $0.50$ & $0.39 \pm 0.32$ & n.a. & n.a. \\
   \midrule
   EndoSurf~\cite{zha_endosurf_2023} & $3.52$ & $2.26 \pm 2.71$ & $8.75$ & $8.19 \pm 3.10$ \\
   \midrule
   ORDT~\cite{hayoz_online_2024} & $3.58$ & $2.45 \pm 2.60$ & $6.22$ & $5.46 \pm 2.98$ \\
   \midrule
   TWPI~\cite{chen_trans-window_2024} & $1.18$ & $0.79 \pm 0.88$ & $3.53$ & $2.98 \pm 1.89$ \\
   \midrule
   TWPI~\cite{chen_trans-window_2024} + tpose & $0.90$ & $0.63 \pm 0.63$ & $0.48$ & $0.40 \pm 0.26$ \\
   \midrule
   Ours & $0.82$ & $0.51 \pm 0.64$ & $3.70$ & $3.16 \pm 1.94$ \\
   \midrule
   Ours + tpose & $0.45$ & $0.37 \pm 0.27$ & $0.44$ & $0.39 \pm 0.21$ \\
   \bottomrule
   \end{tabular}
   \end{center}
   \footnotesize{\textbf{Abbreviations} RMSE: root mean square error; MSD$\pm$std: mean surface distance $\pm$ standard deviation; Non-occluded/Occluded: surface accuracy in the non-occluded and the occluded areas; +tpose: instrument pose-constrained optimization.}
\end{table}

\begin{table}[!t]
   \caption{Geometry recovery accuracy on the DRESDEN4D dataset (unit in millimeter)}
   \label{tab_accuracy2}
   \scriptsize
   \begin{center}
   \begin{tabular}{ccc}
   \toprule
   \multirow{2}{*}{Methods} & \multicolumn{2}{c}{Non-occluded} \\
   & RMSE ($\downarrow$) & MSD$\pm$std ($\downarrow$) \\
   \midrule
   RAFT-Stereo~\cite{lipson_raft-stereo_2021} & $1.88$ & $1.16 \pm 1.48$  \\
   \midrule
   EndoSurf~\cite{zha_endosurf_2023} & n.a. & n.a. \\
   \midrule
   ORDT~\cite{hayoz_online_2024} & $3.52$ & $2.68 \pm 2.28$ \\
   \midrule
   TWPI~\cite{chen_trans-window_2024} & $2.96$ & $2.27 \pm 1.90$ \\
   \midrule
   Ours & $1.98$ & $1.27 \pm 1.52$ \\
   \bottomrule
   \end{tabular}
   \end{center}
   \footnotesize{\textbf{Abbreviations} RMSE: root mean square error; MSD$\pm$std: mean surface distance $\pm$ standard deviation; Non-occluded: surface accuracy in the non-occluded areas.}
\end{table}

\section{Experiment}

\subsection{Implementation details}
The proposed method takes the noisy 3D structure and 3D tracking as the inputs. In this study, the 3D structure is obtained by 3D reconstruction based on a stereo matching method, namely RAFT-Stereo~\cite{lipson_raft-stereo_2021}, and the 3D tracking is obtained in an approach~\cite{chen_occlusion-robust_2023} leveraging a 2D optical flow method, namely LiteFlowNet3~\cite{hui_liteflownet3_2020}. Though not a must, this study leverages the left camera coordinates for defining the image/canonical space. The right view is only used in stereo matching-based depth estimation in the preprocessing. It does not involve tracking and any further processing. We implement the method mainly using MATLAB\textsuperscript{\textregistered} and CUDA\textsuperscript{\textregistered} C++. The method can be run in an online approach with low GPU consumption as it doesn't involve model training. We have tested on two platforms: \textit{Platform 1} CPU AMD\textsuperscript{\textregistered} 5800x with GPU NVIDIA\textsuperscript{\textregistered} 3070, and \textit{Platform 2} CPU Intel\textsuperscript{\textregistered} 13900F with GPU NVIDIA\textsuperscript{\textregistered} 4090. We jointly use MATLAB\textsuperscript{\textregistered} and MeshLab~\cite{cignoni_meshlab_2008} for visualization and rendering.

\subsection{Datasets}
We conducted the experiment using \textit{in vivo} and \textit{ex vivo} binocular laparoscopic datasets. All datasets are publicly available, namely HAMLYN~\cite{mountney_three-dimensional_2010}, ENDONERF~\cite{wang_neural_2022}, SURGEM~\cite{chen_surgem_2024} and DRESDEN4D~\cite{docea_dresden_2026}. Both HAMLYN and ENDONERF are \textit{in vivo}, while SURGEM and DRESDEN4D are \textit{ex vivo}. The cameras in all these datasets were calibrated. The HAMLYN dataset we used involves large camera motion but mild tissue deformation. There is no instrument in the view, and no ground truth 3D structure is provided. ENDONERF involves tool-tissue interactions in the ways of dissection, grasping, and traction, where the camera is always fixed. Instrument masks are given to indicate the occluded area. No ground truth geometry is provided. SURGEM involves tool-tissue interaction in the way of palpation, where the camera is also fixed. Instrument masks and poses and ground truth 3D structures are provided. The ground truth 3D structures are acquired by 3D scanning from the back of thin tissue; thus, the structure in both the non-occluded and occluded areas is provided. For DRESDEN4D, we utilize their \textit{moving-camera-and-static-tissue} data, where the ground truth 3D structure of the scene is provided by 3D scanning.

\subsection{Experiment methods and evaluation metrics}
In section~\ref{sec_result_3d}, we quantitatively evaluate the 3D reconstruction accuracy with either \textit{static-camera-and-deforming-tissue} or \textit{moving-camera-and-static-tissue} settings using the SURGEM and DRESDEN4D datasets, respectively. Root mean square error (RMSE) and mean surface distance (MSD)~\cite{chen_occlusion-robust_2023} are used to quantify the 3D structure accuracy by comparing the reconstructed and ground truth 3D geometry. The proposed method is compared to the previous tissue deformation recovery methods. For fairness, we only compare those methods that have guaranteed geometric accuracy, are declared to be robust to occlusion, and are designed for deformable scene, including EndoSurf~\cite{zha_endosurf_2023}, TWPI~\cite{chen_trans-window_2024}, and ORDT~\cite{hayoz_online_2024}. All implemented methods use RAFT-Stereo~\cite{lipson_raft-stereo_2021} to acquire 3D reconstructed scenes as input. The RAFT-Stereo-derived 3D data is also used for reference, which is not robust to occlusion. Note that all comparison methods are camera-pose-based, where EndoSurf and TWPI estimate the pose internally and ORDT requires external input of the pose. An ablation study was conducted to further demonstrate that the proposed method works on noisy inputs and is robust to accumulative drift, where the deformation optimization module was disabled and the error from visual deformation measurement was directly passed into the final output, resulting in foreseeable larger accumulative errors.
Due to the difficulty in ground truth geometry acquisition, the reconstruction accuracy under the \textit{moving-camera-and-deforming-tissue} setting is only qualitatively evaluated using various datasets in section~\ref{sec_result_more}.

In section~\ref{sec_result_revisit}, experiments were conducted using the HAMLYN dataset to quantify the revisit error and long-term stability of the proposed method, where the camera was always moving. We generated forward-reverse videos by appending a reversed sequence to the original clip for the evaluation~\cite{recasens_drunkards_2023}. The frame where the video reverses varies; the further this point is from the start, the longer the video becomes, leading to a potentially higher revisit error. RMSE and MSD are used to measure the difference between the reconstructed structures of the final and the first frame, which are ideally identical.

Section~\ref{sec_result_more} shows more qualitative deformation recovery results in various surgical manipulations to further demonstrate the generalizability of the proposed method. Surface strain estimation, as a natural byproduct of the proposed deformation representation, is only qualitatively evaluated in section~\ref{sec_result_more}.

Finally, runtime performance of the proposed method is evaluated in section~\ref{sec_result_runtime}.

\begin{figure*}[!t]%
\centering
\includegraphics[width=0.8\textwidth]{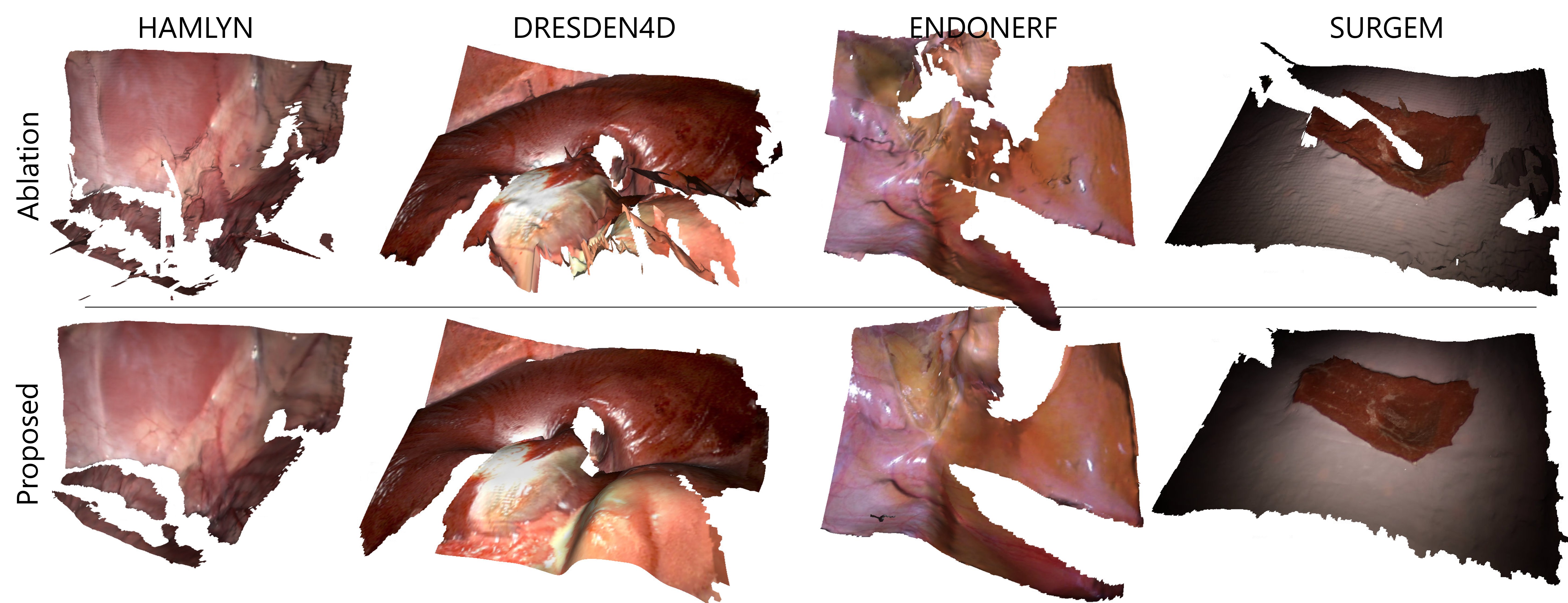}
\caption{Reconstructed tissue in the ablation study. Results in the first row were acquired when the deformation optimization was disabled.} \label{fig_ablation}
\end{figure*}

\section{Result and discussion}

\subsection{3D reconstruction accuracy}\label{sec_result_3d}

Quantitative experiments on the SURGEM dataset~\cite{chen_surgem_2024} were conducted to evaluate the 3D reconstruction accuracy under the \textit{fixed-camera-and-deforming-tissue} case in both the non-occluded and the occluded areas. Results are shown in Table~\ref{tab_accuracy}. All the methods involved in this experiment rely on the RAFT-Stereo result as their input, and thus, the RAFT-Stereo accuracy can be regarded as a reference. However, since RAFT-Stereo is not robust to occlusion, no information can be acquired in the occluded area. As can be seen from Table~\ref{tab_accuracy}, in the case when instrument pose information is available, the proposed method shows the highest accuracy in both the non-occluded and the occluded areas. In the case when tool-induced occlusion occurs but instrument pose information is unavailable, the accuracy of the proposed method drops in the occluded area, which is a foreseeable result. Besides, we want to highlight that the proposed method, as an online method, outperforms EndoSurf~\cite{zha_endosurf_2023}, which is offline, showing promising potential for intraoperative applications. As a training-free online method, the proposed method also outperforms another training-based online method, ORDT~\cite{hayoz_online_2024}, which performs iterative training for 3D gaussian in each new frame. Note that the evaluation result is affected by the inevitable registration error and bias inherited from the SURGEM dataset. Thus, Hotelling's T-squared test was conducted for significance analysis~\cite{hotelling_generalization_1992}. Results show that the proposed method (with pose-constrained optimization) significantly outperforms all methods in both the occluded and non-occluded areas (significant with p-value smaller than 0.05), while the only one exception is in the occluded area as compared to~\cite{chen_trans-window_2024} (insignificant with p-value larger than 0.05). Results of significance analysis prove that even given enhanced flexibility of modeling deformation and not using camera pose, the method can still achieve significantly higher or at least same level of accuracy as compared to previous methods.

Experiments on the DRESDEN4D~\cite{docea_dresden_2026} dataset were for evaluating under the \textit{moving-camera-and-static-tissue} case in only the non-occluded area. Similarly, the RAFT-Stereo accuracy is for reference. Results in Table~\ref{tab_accuracy2} show that accurate reconstruction can be achieved even under severe camera motion and that the proposed method outperforms others. Note that EndoSurf failed to generate plausible results and is marked as n.a. in Table~\ref{tab_accuracy2}.

Note that all comparison methods are camera-pose-based. In the \textit{fixed-camera-and-deforming-tissue} case (SURGEM), camera poses in all frames can be regarded as identical and thus can be regarded as noise-free. While in the \textit{moving-camera-and-static-tissue} case (DRESDEN4D), all comparison methods need to estimate camera pose, and thus the associated error exists. Bad camera pose estimation leads to misalignment of inter-frame data, resulting in structure diffusion. The proposed method, on the other hand, eliminates the camera pose-derived error from the root and achieves the state-of-the-art performance. This makes us believe that camera pose may not be necessary for the reconstruction of dynamic scenes under the assumption that part of the visible tissue is always connected to the invisible one.

\textbf{Ablation study:} Results were acquired when the deformation optimization module was disabled. The noises from the visual measurement accumulated and resulted in surface discontinuity and roughness, as shown in Fig.~\ref{fig_ablation}. In the SURGEM dataset, reconstruction error (RMSE) significantly increases from 0.82 mm to 16.81 mm and from 3.70 mm to 31.74 mm in the non-occluded and the occluded areas, respectively. In the DRESDEN4D dataset, reconstruction error (RMSE) significantly increases from 1.98 mm to 57.12 mm. Results prove that the proposed method works on noisy inputs and is robust to accumulative drift.

\begin{figure}[!t]%
\centering
\includegraphics[width=0.49\textwidth]{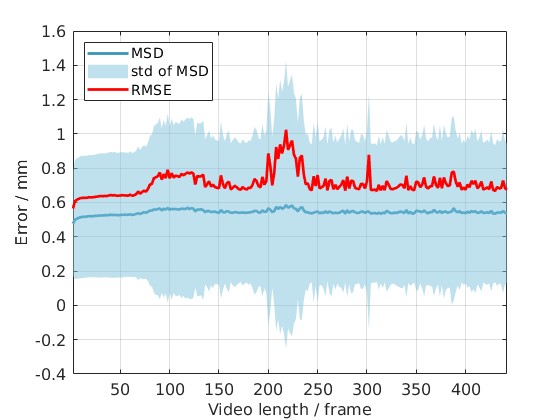}
\caption{The video length vs. revisit error.} \label{fig_revisit}
\end{figure}

\subsection{Revisit accuracy}\label{sec_result_revisit}

Results in Fig.~\ref{fig_revisit} show the tendency of the revisit error with respect to the increase of video frames. The revisit error (RMSE) increased from 0.6 mm to 0.8 mm when the number of frames gradually increased to 100 but later became stable at around 0.7 mm. The std of MSD indicates that errors fluctuate in around 0.4 mm.
Given that the spatial resolution of the reconstructed tissue is around 0.4 mm, such fluctuation and average revisit error are not significant. Results prove that the proposed method has stable revisit performance. Benefiting from the camera-centric setting, the proposed method does not face significant coordinate drift in the first and the last frames in the experiment using forward-reverse videos.

\subsection{Deformation recovery and surface strain estimation in various surgical manipulation}\label{sec_result_more}

This section presents the recovery results of tissue deformation in various surgical manipulations, including camera motion, traction, dissection, and palpation. In addition, the results of the surface strain estimation are also presented in the cases of traction and palpation, where the tissue deformation is continuous and significant. 

\begin{figure*}[!t]%
\centering
\includegraphics[width=1\textwidth]{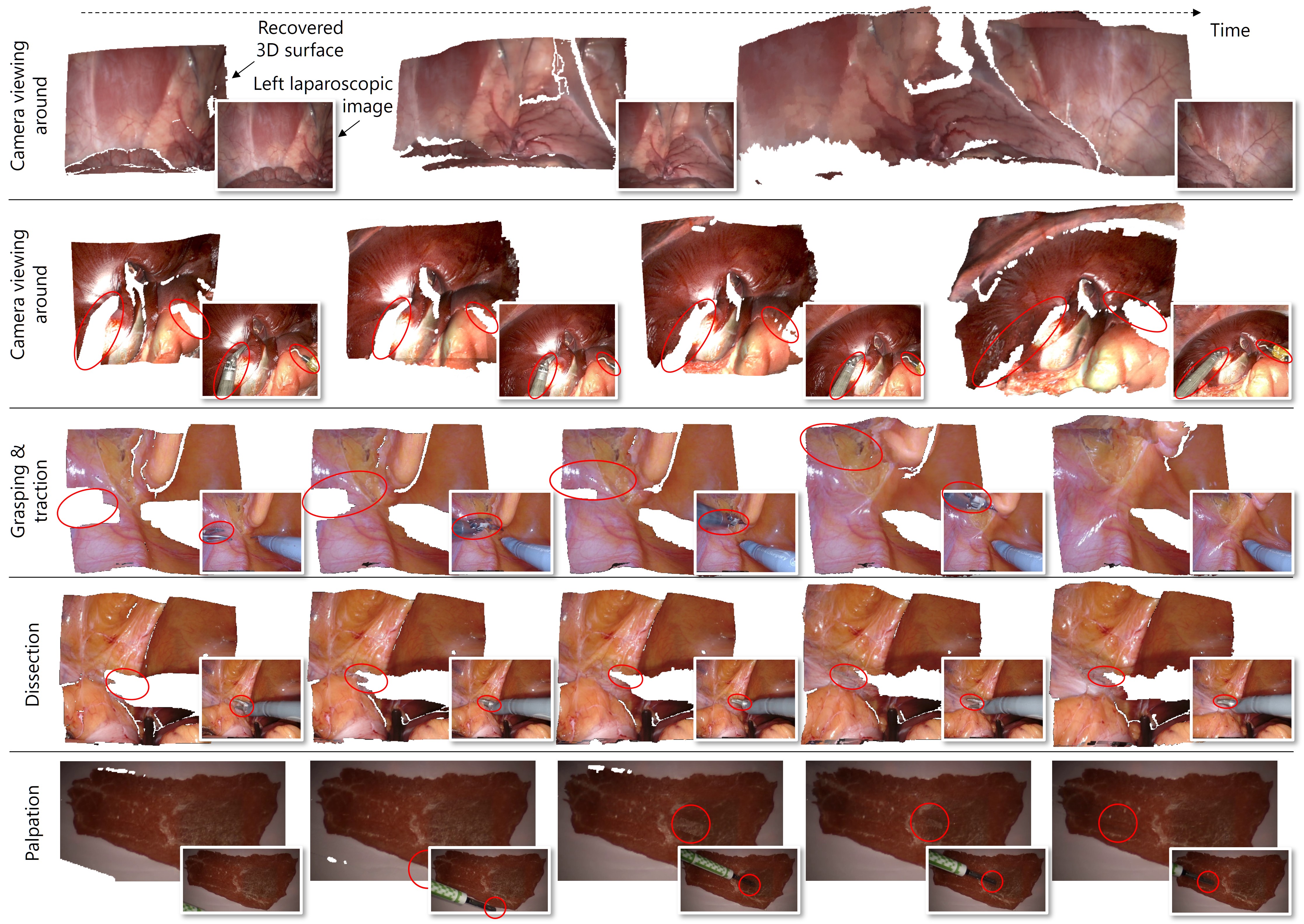}
\caption{Results of the proposed online deformation recovery method together with the corresponding laparoscopic image frames. Time increases from the left to the right. Red rings indicate the locations of the instrument. In the first and the second row, the cameras are moving around viewing the abdominal tissues, but the tissue deformation is mild (the first row) or none (the second row).} In the third row, two forceps cause severe occlusion. One forceps perform grasping and traction, while the other is idle. In the fourth row, one forceps is performing dissection, leading to the change in tissue surface continuity. In the fifth row, one forceps approach and push the tissue surface, and then leave. \label{fig_case}
\end{figure*}

Fig.~\ref{fig_case} shows the online behavior of the proposed method. The method gradually recovers the structure of a deformable scene. The structure that is initially occluded by the instrument or is outside the field of view will be added once it becomes visible. On the other hand, if the structure used to be observable but is currently occluded or outside the field of view, it will still be maintained in the recovered results, proving that the proposed method is robust to occlusion and camera motion. When the geometric continuity of the tissue changes in the case of dissection, the proposed method can model the newly exposed surface and update the geometry and texture to make it consistent with the latest observation. However, the proposed method does not tend to recover the permanently invisible structure, leaving some vacant regions, as can be seen in Fig.~\ref{fig_case}.

\begin{figure*}[!t]%
\centering
\includegraphics[width=0.9\textwidth]{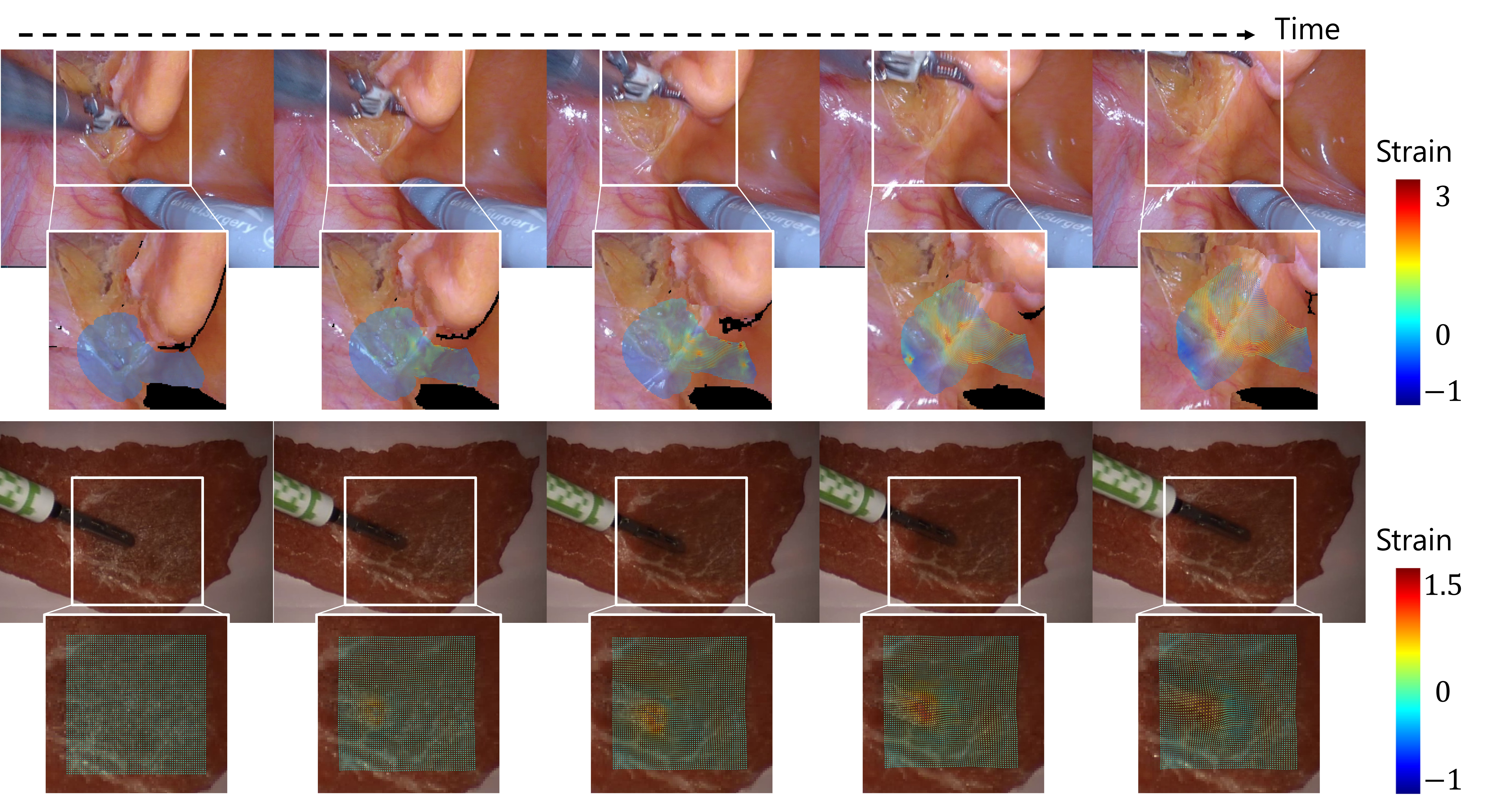}
\caption{The estimated surface maximal strain using the proposed method. Time increases from the left to the right. The first and the third rows are the images from the laparoscopic camera. The second and the fourth rows are the colored maximal surface principal strain overlaid with the recovered tissue surface. The first and the second rows show the case in traction, where the tissue surface is stretched. The third and the fourth rows show the case in palpation, where the forceps approach, push, and leave the tissue.} \label{fig_strain}
\end{figure*}

Fig.~\ref{fig_strain} shows the estimated surface strain overlaid on the recovered tissue structure, together with the corresponding laparoscopic frame image. We want to clarify that the surface strain visualized in Fig.~\ref{fig_strain} is the maximal principal inner-surface small strain. In the previous method~\cite{chen_trans-window_2024}, surface strain estimation is not convenient since only the displacement map is modeled, while the proposed method directly models the stretching effect in the deformation map, from which surface strain can be easily obtained. We manually select the region of interest to continuously track surface points and measure the strain, as shown in Fig.~\ref{fig_strain}. The surface strain is initially zero and gradually increases when tissue deformation becomes severe. The surface strain distribution is inhomogeneous as the deformation is non-affine. In the case of traction, the maximal strain occurs around the center line of the stretched surface, while in the case of palpation, the maximal strain occurs around the forceps tip. Results show that the proposed method can estimate surface strain under different manipulations with robustness to occlusion, and that the laparoscopic image-derived surface strain may be a new and useful modality for tool-tissue interaction analysis.

\begin{figure}[!t]%
\centering
\includegraphics[width=0.47\textwidth]{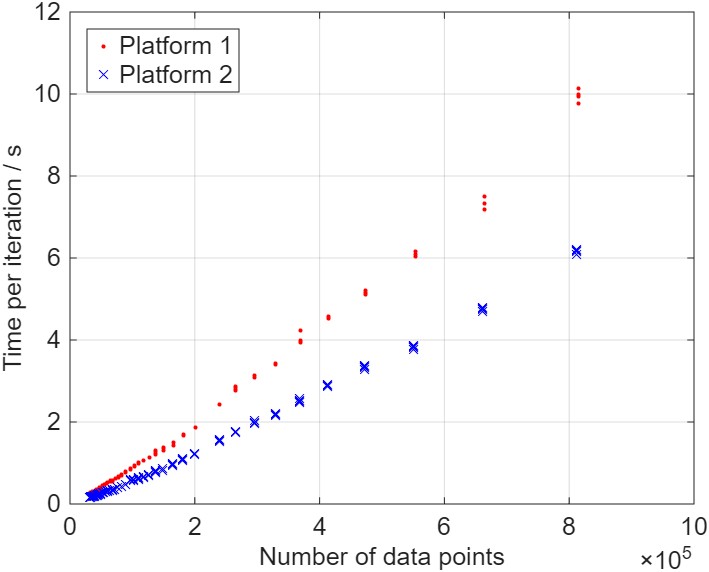}
\caption{Runtime performance of the proposed deformation optimization. \textit{Platform 1}: CPU AMD\textsuperscript{\textregistered} 5800x with GPU NVIDIA\textsuperscript{\textregistered} 3070. \textit{Platform 2}: CPU Intel\textsuperscript{\textregistered} 13900F with GPU NVIDIA\textsuperscript{\textregistered} 4090.} \label{fig_runtime}
\end{figure}

\begin{table*}[!t]
   \caption{Trade-off analysis of downsampling and accuracy.}
   \label{tab_sr_acc}
   \scriptsize
   \begin{center}
   \begin{tabular}{cccccc}
   \toprule
   Number of data points ($\times 10^5$) & $8.15$ & $2.02$ & $0.90$ & $0.49$ & $0.32$ \\
   \midrule
   Median distance to neighbor points (mm) & $0.20$ & $0.40$ & $0.59$ & $0.79$ & $0.98$ \\
   \midrule
   Reconstruction error in RMSE (mm) & $0.70$ & $0.74$ & $0.76$ & $0.81$ & $0.87$ \\
   \midrule
   p-value & n.a. & $0.78$ & $0.24$ & $0.01$ & $<0.001$ \\
   \bottomrule
   \end{tabular}
   \end{center}
   \footnotesize{RMSE: root mean square error.}
\end{table*}

\subsection{Runtime performance}\label{sec_result_runtime}
As an online method, we evaluate the runtime performance of the proposed method by modifying the number of data points and measuring the time per iteration. Fig.~\ref{fig_runtime} shows the time cost of deformation optimization, which is the core of the proposed method. Given an image with the resolution of $W\times H$, if there is no downsampling or compressing process, the initial number of data points will be the same as that of the pixels and will gradually increase if the field of view changes. Take an image in the SURGEM dataset as an example; it has the resolution of $1280\times720$, resulting in the initial number of data points of $921600$ and around $8.15\times 10^5$ after outlier detection. It takes about $9.96$ seconds for an iteration on \textit{Platform 1} and $6.16$ seconds on \textit{Platform 2}. If we downsample the data points from $8.15\times 10^5$ to $3.2 \times 10^4$, the time per iteration drops to $0.25$ second on \textit{Platform 1} and $0.16$ second on \textit{Platform 2}. As shown in Fig.~\ref{fig_runtime}, the time per iteration is almost proportional to the number of data points. This confirms that the computational cost empirically scales linearly with the number of data points, which aligns directly with the theoretical runtime trends discussed in the complexity analysis in section~\ref{sec_optimization}. We believe that the current bottleneck of the runtime performance is in the CPU side, as the solver of the optimization problem is working on CPU only, and the GPU consumption is always low (less than $10\%$) during the optimization. For complete runtime testing please refer to appendix~\ref{appendix_runtime}.

Trade-off analysis of downsampling and deformation recovery accuracy is conducted, as shown in Table~\ref{tab_sr_acc}. The number of data points indicates the downsampling rate, where $8.15\times 10^5$ is the amount of data that is close to lossless data with image resolution of $1280\times720$, thus, the result of which is used as reference when conducting Hotelling’s T-squared test~\cite{hotelling_generalization_1992}. Results show that there is no significant difference in the results with number of downsampled data points of $2.02\times 10^5$ and $0.90\times 10^5$ at the 5\% significance level as compared to the reference one, proving that downsampling does not significantly deteriorate the deformation recovery accuracy. However, with the increase of downsampling rate (as for data points of $0.49\times 10^5$ and $0.32\times 10^5$), median distance to neighbor points shows the downgrade in the spatial resolution of the recovered geometry. Hotelling’s T-squared test and accuracy results also prove that there is significant deterioration in the reconstruction accuracy. The trade-off between geometric complexity and runtime performance needs to be considered based on the actual application situation. For our experiments, we suggest downsample data points into $9.0\times 10^4$ for better runtime performance with insignificant accuracy deterioration.

\subsection{Limitations}

Though continuous deformable scene reconstruction when the camera is moving has been demonstrated in section~\ref{sec_result_more}, the proposed method does not optimize the stretching effect for those geometries outside the field of view. Instead, only the 3D position and orientation of these geometries are optimized. In other words, the geometry outside the field of view is regarded as a rigid body connecting with the deformable geometry inside the field of view. The biomechanical property and boundary condition of some of the invisible tissue may be different from those of the visible one, making it less reliable to estimate the deformation of the invisible tissue that locates far away from the visible one. Moreover, projection-based parameterization is implemented in this article, which cannot model self-folding geometry nor the geometry at the backside of the camera. In the context of clinical applicability, this restricts the current method from being used in unconstrained bowel manipulation or fully collapsed cavities.

Since the proposed method only utilizes inter-frame data for optimization and fusion, long-term stability is not guaranteed. Misalignment may occur during a revisit, and the current method simply replaces the historical data with the latest one. However, smoothing using historical data based on concepts like bundle adjustment~\cite{guo_cmax-slam_2024} may help to improve the long-term stability of frame alignment. Additionally, as a camera-pose-free approach, the proposed method aligns frame data based on the assumption of tissue connectivity. Therefore, when the tissue connection becomes weak, the estimation becomes unstable.

The proposed method requires absolute depth as input. Though it does not theoretically require the use of a binocular vision system, the current relative-depth-only technique in monocular vision limits its application in scenarios such as endoluminal surgery, where  a monocular endoscopic camera is commonly used.

Realtime performance is another concern when trying to yield an method for intraoperative application. The proposed method has online behavior, but we still carefully choose to not use the word ``realtime'' at the current stage. Although in the experiment we can see that the processing time of per-frame data can be lower than 0.2 second, this does not count the time for acquiring the initial depth and optical flow data, which varies depending on the exact method in use and the data resolution. Besides the data acquisition time, the current bottleneck remains in the CPU code for solving the constrained linear programming problem.

\section{Conclusion}

This article presents a vision-based method for continuous tissue deformation recovery. Equipped with the noval camera-centric setting and high-dimensional mapping-based deformation optimization, it fully eliminates the use of camera pose and demonstrates its robustness under dynamic and deformable intraoperative environments, even with occlusion and irregular camera/tissue motion. With the tissue-invariant vision-biomechanics consistency constraints, the proposed method can be adapted to various tissues under various surgical manipulations with no effort. The proposed method also helps to estimate and visualize the intraoperative surface strain during soft tissue manipulation. We believe the surface strain can be regarded as a new modality that directly correlates to tissue biomechanical behaviors and benefits many downstream tasks like tool-tissue interaction analysis, risk assessment, and surgical task automation.

\section*{Acknowledgments}
This research was supported by the JST Moonshot R\&D Grant Number JPMJMS2214, the Grant-in-Aid for JSPS Fellows (Number 25KJ1017), and the Kayamori Foundation of Informational Science Advancement.

{\appendices

\section{Algorithm pseudocode}\label{appendix_pseudo}

Please see Algorithm~\ref{al_pre},~\ref{al_opti},~\ref{al_reparam}, and~\ref{al_fuse}.

\begin{algorithm}[h]
\label{al_pre}
    \scriptsize
	\SetKwData{Left}{left}
	\SetKwData{This}{this}
	\SetKwData{Up}{up}
	\SetKwFunction{Union}{Union}
	\SetKwFunction{FindCompress}{FindCompress}
	\SetKwInOut{Input}{Input}
	\SetKwInOut{Output}{Output}
    
    \Input{Binocular images at frame $t$ and $t+1$}
	\Output{Geometry maps $\bm{\mathcal{P}}_I|_t$, $d\bm{\mathcal{P}}_I|_t$, $\bm{\mathcal{P}}_I|_{t+1}$, $d\bm{\mathcal{P}}_I|_{t+1}$, deformation maps $\bm{\mathcal{F}}_I|_t$, $\bm{\mathcal{D}}_I|_t$, and parameter sets $U_I|_t$ and $U_I|_{t+1}$}
	\BlankLine
   
	Depth estimation using binocular images of frame $t$ and $t+1$, parameterize depth maps into 3D point maps $\bm{\mathcal{P}}_I|_t$ and $\bm{\mathcal{P}}_I|_{t+1}$ with the associated parameter sets $U_I|_t$ and $U_I|_{t+1}$, respectively\;
    Calculate the derivatives of the 3D point map and obtain $d\bm{\mathcal{P}}_I|_t$ and $d\bm{\mathcal{P}}_I|_{t+1}$\;
    Calculate the 2D optical flow using only the left images of frame $t$ and $t+1$ and obtain $\bm{\mathcal{O}}_I|_t$\;
    Combine 3D optical flow map and geometry maps to calculate raw inter-frame deformation $\bm{\mathcal{F}}_I|_t$ and $\bm{\mathcal{D}}_I|_{t}$.

	\caption{Preprocessing (parameterization)}
\end{algorithm}

\begin{algorithm}[h]
\label{al_opti}
    \scriptsize
	\SetKwData{Left}{left}
	\SetKwData{This}{this}
	\SetKwData{Up}{up}
	\SetKwFunction{Union}{Union}
	\SetKwFunction{FindCompress}{FindCompress}
	\SetKwInOut{Input}{Input}
	\SetKwInOut{Output}{Output}
    
    \Input{Geometry maps $\bm{\mathcal{P}}_C|_t$, $d\bm{\mathcal{P}}_C|_t$ and their associated parameter set $U_C|_t$, and deformation maps $\bm{\mathcal{F}}_I|_t$, $\bm{\mathcal{D}}_I|_t$ and their associated parameter set $U_I|_t$}
	\Output{Deformation maps $\widehat{\bm{\mathcal{F}}}_C|_t$, $\widehat{\bm{\mathcal{D}}}_C|_t$}
	\BlankLine
   
	Detect outliers and separate $U_C|_t$ into the inlier set $U_{C,i}|_t$ and the outlier set $U_{C,o}|_t$\;
    Optimize $\bm{\mathcal{D}}_C|_t$ using $\bm{\mathcal{D}}_I|_t$; the estimated result is $\widehat{\bm{\mathcal{D}}}_C|_t$\;
    Deform the local geometry $d\bm{\mathcal{P}}_C|_t$ using $\widehat{\bm{\mathcal{D}}}_C|_t$ and get $d\widehat{\bm{\mathcal{P}}}'_C|_t$\;
    \uIf{\textup{surgical instrument pose is available}}{Optimize $\bm{\mathcal{F}}_C|_t$ using $\bm{\mathcal{F}}_I|_t$, $\bm{\mathcal{P}}_C|_t$, and $d\widehat{\bm{\mathcal{P}}}'_C|_t$ subject to the instrument pose constraint; the estimated result is $\widehat{\bm{\mathcal{F}}}_C|_t$\;}
	\Else{
    Optimize $\bm{\mathcal{F}}_C|_t$ using $\bm{\mathcal{F}}_I|_t$, $\bm{\mathcal{P}}_C|_t$, and $d\widehat{\bm{\mathcal{P}}}'_C|_t$; the estimated result is $\widehat{\bm{\mathcal{F}}}_C|_t$.}

	\caption{Optimization of the canonical deformation}
\end{algorithm}

\begin{algorithm}[h]
\label{al_reparam}
    \scriptsize
	\SetKwData{Left}{left}
	\SetKwData{This}{this}
	\SetKwData{Up}{up}
	\SetKwFunction{Union}{Union}
	\SetKwFunction{FindCompress}{FindCompress}
	\SetKwInOut{Input}{Input}
	\SetKwInOut{Output}{Output}
    
    \Input{Geometry maps $\bm{\mathcal{P}}_C|_t$, $d\bm{\mathcal{P}}_C|_t$, deformation maps $\widehat{\bm{\mathcal{F}}}_C|_t$, $\widehat{\bm{\mathcal{D}}}_C|_t$, and texture map $\bm{\mathcal{C}}_C|_t$}
	\Output{Geometry maps $\widehat{\bm{\mathcal{P}}}'_C|_{t+1}$, $d\widehat{\bm{\mathcal{P}}}'_C|_{t+1}$, texture map $\widehat{\bm{\mathcal{C}}}'_C|_{t+1}$, and their associated parameter set $\widehat{U}'_C|_{t+1}$}
	\BlankLine
   
	Deform the geometry maps $\bm{\mathcal{P}}_C|_t$ and $d\bm{\mathcal{P}}_C|_t$ using $\widehat{\bm{\mathcal{F}}}_C|_t$ and $\widehat{\bm{\mathcal{D}}}_C|_t$, and get $\widehat{\bm{\mathcal{P}}}'_C|_t$ and $d\widehat{\bm{\mathcal{P}}}'_C|_t$ with the associated geometry $\widehat{S}'_C|_{t}$ and $d\widehat{S}'_C|_{t}$, respectively\;
    Parameterize $\widehat{S}'_C|_{t}$ and $d\widehat{S}'_C|_{t}$ and get $\widehat{\bm{\mathcal{P}}}'_C|_{t+1}$ and $d\widehat{\bm{\mathcal{P}}}'_C|_{t+1}$ with the associated parameter set $\widehat{U}'_C|_{t+1}$, respectively\;
    Update the bound texture from $\bm{\mathcal{C}}_C|_{t}$ to $\widehat{\bm{\mathcal{C}}}'_C|_{t+1}$.

	\caption{Reparameterization}
\end{algorithm}

\begin{algorithm}[h]
\label{al_fuse}
    \scriptsize
	\SetKwData{Left}{left}
	\SetKwData{This}{this}
	\SetKwData{Up}{up}
	\SetKwFunction{Union}{Union}
	\SetKwFunction{FindCompress}{FindCompress}
	\SetKwInOut{Input}{Input}
	\SetKwInOut{Output}{Output}
    
    \Input{Geometry maps $\bm{\mathcal{P}}_I|_{t+1}$, $d\bm{\mathcal{P}}_I|_{t+1}$, texture map $\bm{\mathcal{C}}_I|_{t+1}$, and their associated parameter set $U_I|_{t+1}$; geometry maps $\widehat{\bm{\mathcal{P}}}'_C|_{t+1}$, $d\widehat{\bm{\mathcal{P}}}'_C|_{t+1}$, texture map $\widehat{\bm{\mathcal{C}}}'_C|_{t+1}$, and their associated parameter set $\widehat{U}'_C|_{t+1}$}
	\Output{Geometry maps $\widehat{\bm{\mathcal{P}}}_C|_{t+1}$, $d\widehat{\bm{\mathcal{P}}}_C|_{t+1}$, texture map $\widehat{\bm{\mathcal{C}}}_C|_{t+1}$, and their associated parameter sets $\widehat{U}_C|_{t+1}$}
	\BlankLine
   
	Fuse the parameter set $\widehat{U}_C|_{t+1}=\widehat{U}'_C|_{t+1}\cup U_I|_{t+1}$\;
    Initialize $\widehat{\bm{\mathcal{P}}}_C|_{t+1}$, $d\widehat{\bm{\mathcal{P}}}_C|_{t+1}$ and $\widehat{\bm{\mathcal{C}}}_C|_{t+1}$ \;
    \For {\textup{each data point }$(u,v)\in U_C|_{t+1}$} {\uIf{\textup{is texture map}}{\uIf{\textup{there is data at data point }$(u,v)$\textup{ in the image space}}{keep the data in the image space}\Else{keep the data in the canonical space}}\Else{\uIf{\textup{there is data at data point }$(u,v)$\textup{ in the canonical space}}{keep the data in the canonical space}\Else{keep the data in the image space}}}

	\caption{Map fusion}
\end{algorithm}

\section{Inter-frame strain threshold}\label{appendix_strain}

This is a high-bound threshold to filter out the abnormal strain signal from the measurement, under the assumption that the surgical instrument and the tissue will not move or deform fast. Let's note the threshold as $\varepsilon_t$, the laparoscopic/endoscopic video stream frame rate as $f_s$, the robot end-effector speed as $v$, and the tissue original size as $l_0$. The threshold can be estimated by:

\begin{equation}
    \varepsilon_t=\frac{v}{f_sl_0}
\end{equation}

In this study, a common laparoscopic video stream has a frame rate of 30 FPS (frame per second) and the field of view contains tissue with the size of around 5 cm. The robot end-effector speed is unlikely to exceed 15 cm/s. Thus, the threshold is set to 0.1 for all experiments.

\section{Runtime performance}\label{appendix_runtime}

The runtime of processing depth, optical flow, data acquisition, and major components of the proposed method was tested and reported in Table~\ref{tab_app_runtime}. The data was acquired using the SURGEM dataset with the resolution of $1280\times720$ and a number of $0.90\times 10^5$ sampled data points. Note that the proposal is compatible with other depth and optical flow estimation methods, and users are recommended to replace them with more efficient ones if conditions permit.

\begin{table}[!t]
   \caption{Runtime performance.}
   \label{tab_app_runtime}
   \scriptsize
   \begin{center}
   \begin{tabular}{lcc}
   \toprule
   Preprocessing & \textit{Platform 1} & \textit{Platform 2} \\
   \midrule
   ~~~~3D reconstruction (stereo matching)~\cite{lipson_raft-stereo_2021} & 2.1683 s & 0.7442 s  \\
   \midrule
   ~~~~2D tracking (optical flow)~\cite{hui_liteflownet3_2020} & 1.3635 s & 0.9644 s \\
   \midrule
   The proposed & & \\
   \midrule
   ~~~~Deformation optimization & 0.8857 s & 0.4832 s \\
   \midrule
   ~~~~Reparameterization & 0.1732 s & 0.1527 s \\
   \midrule
   ~~~~Map fusion & 0.0042 s & 0.0050 s \\
   \bottomrule
   \end{tabular}
   \end{center}
   \footnotesize{\textit{Platform 1}: CPU AMD\textsuperscript{\textregistered} 5800x with GPU NVIDIA\textsuperscript{\textregistered} 3070. \textit{Platform 2}: CPU Intel\textsuperscript{\textregistered} 13900F with GPU NVIDIA\textsuperscript{\textregistered} 4090.}
\end{table}

}



\bibliographystyle{IEEEtran}
\bibliography{reference}

\vspace{11pt}
\vspace{-33pt}
\begin{IEEEbiography}[{\includegraphics[width=1in,height=1.25in,clip,keepaspectratio]{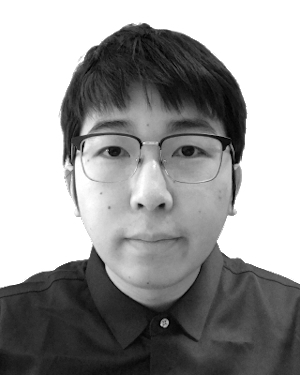}}]{Jiahe Chen} received the Ph.D. in Precision Engineering and M.S. in Bioengineering from the University of Tokyo, Tokyo, Japan, in 2022 and 2025, respectively; and B.S. in Mechanical Engineering from Beihang University, Beijing, China, in 2020. He was a JSPS research fellow in the department of Precision Engineering, School of Engineering, The University of Tokyo, from 2025 to 2026. His research interest lies in the area of computer assisted intervention, surgical robotics (perception and control), and surgical simulation.
\end{IEEEbiography}

\vspace{11pt}

\vspace{-33pt}
\begin{IEEEbiography}[{\includegraphics[width=1in,height=1.25in,clip,keepaspectratio]{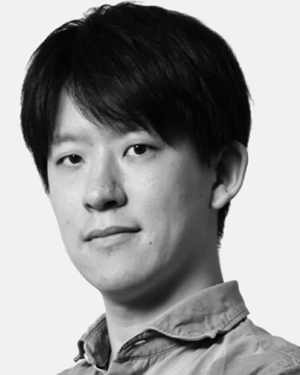}}]{Naoki Tomii} received Ph.D. degree in engineering from The University of Tokyo, Tokyo, Japan, in 2017. He is currently an Associate Professor in the Research Center for Advanced Science and Technology, the University of Tokyo. His current research interests include signal processing, AI, and computer simulation in medicine. He is currently in charge of several research grants from Japan in Biomedical Engineering. He belongs to IEEE, Japanese Society for Medical and Biological Engineering (JSMBE), Japanese Society for Computer Aided Surgery, Japanese Heart Rhythm Society.
\end{IEEEbiography}

\vspace{11pt}

\vspace{-33pt}
\begin{IEEEbiography}[{\includegraphics[width=1in,height=1.25in,clip,keepaspectratio]{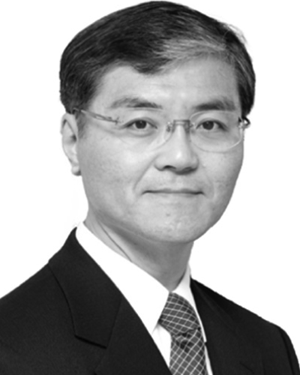}}]{Ichiro Sakuma} received the B.S., M.S., Ph.D. in Precision Engineering from The University of Tokyo, in 1982, 1984, and 1989 respectively. He was Research Associate from 1985 to 1987 in Department Precision Engineering in Department of Precision Engineering, The University of Tokyo. He was Research Associate, a Lecturer, and Associate Professor in School of Science and Engineering, Tokyo Denki University from 1987 to 1998. He was associate Professor from 1998 to 1999 in School of Engineering, Associate Professor from 1999 to 2001, and Professor from 2001 to 2006 in Graduate School of Frontier Sciences, The University of Tokyo. He was a Professor in School of Engineering, The University of Tokyo, from 2006 to 2025. He was the Vice Dean of School of Engineering from 2014 to 2017. He has been the director of Research Institute for Biomedical Science and Engineering in the same university. He has been a Emeritus Professor of The University of Tokyo and a Distinguished Professor in the Research Institute for Science and Technology, Center for Research and Collaboration, Tokyo Denki University, since 2025. 
His research interests include biomedical instrumentation, cardiac arrhythmias, computer-assisted intervention, and surgical robotics. He is a fellow of International Academy of Medical and Biological Engineering (IAMBE). He is a Board Member of Japanese Society for Medical and Biological Engineering (JSMBE), and President of JSMBE from 2015 to 2016. He is an editorial board member of IEEE Transactions on Biomedical Engineering. He is a Board member of Japan Society of Computer Aided Surgery, International Society for Computer Aided Surgery, Japanese Heart Rhythm Society.
\end{IEEEbiography}

\vspace{11pt}

\vspace{-33pt}
\begin{IEEEbiography}[{\includegraphics[width=1in,height=1.25in,clip,keepaspectratio]{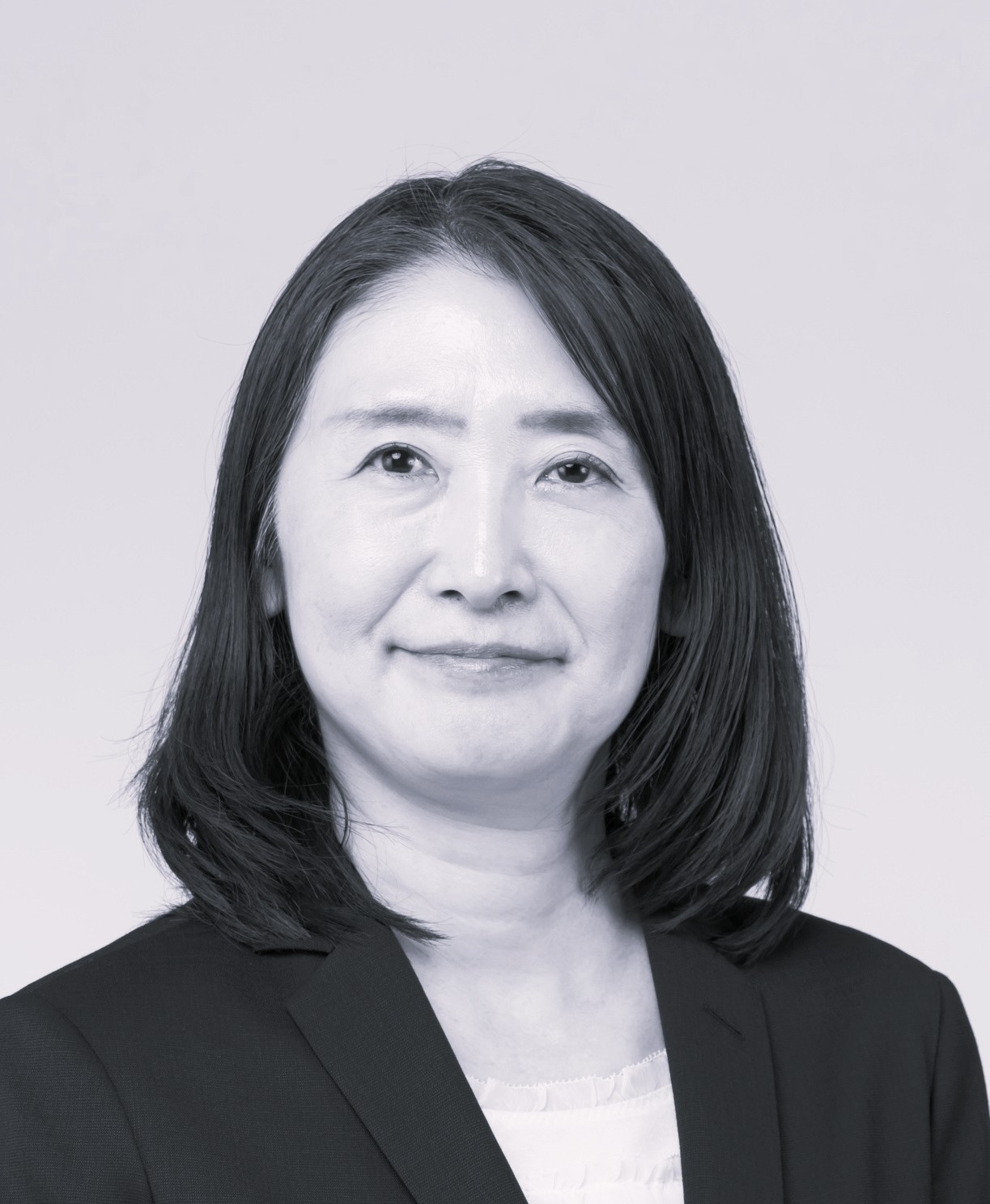}}]{Etsuko Kobayashi} received the B.S., M.S., and Ph.D. degrees in precision machinery engineering from the University of Tokyo, Tokyo, Japan, in 1995, 1997, and 2000, respectively. From 2000 to 2003, she was a Research Associate in the School of Frontier Science, University of Tokyo. From 2003 to 2006, she was a Lecturer of the School of Frontier Science, University of Tokyo. From 2006 to 2017, she was an Associate Professor in the Department of Precision Engineering, School of Engineering, University of Tokyo. From 2017 to 2019, she was an Associate Professor in the School of Medicine, the Tokyo Women's Medical University. From 2020, she is with the Department of Precision Engineering, School of Engineering, University of Tokyo, where she is currently a full Professor.
Her research interests include medical robotics, surgical navigation systems, and biomedical instrumentation. Prof. Kobayashi is a member of the International Society for Computer Aided Surgery, the Japan Society of Computer Aided Surgery, and the Japanese Society for Medical and Biological Engineering.
\end{IEEEbiography}

\vfill

\end{document}